\documentclass[journal]{IEEEtran}
\usepackage{amsmath}
\usepackage{amssymb}
\usepackage{booktabs}
\usepackage{color}
\usepackage{cite}

\ifCLASSINFOpdf
   \usepackage[pdftex]{graphicx}
   \graphicspath{{./figs/}}
   \DeclareGraphicsExtensions{.pdf,.jpeg,.png}
\else
   \usepackage[dvips]{graphicx}
   \graphicspath{{./figs/}}
   \DeclareGraphicsExtensions{.eps}
\fi

\usepackage{algorithm}
\usepackage{algorithmic}

\usepackage{array}

\ifCLASSOPTIONcompsoc
  \usepackage[caption=false,font=normalsize,labelfont=sf,textfont=sf]{subfig}
\else
  \usepackage[caption=false,font=footnotesize]{subfig}
\fi

\begin{document}

\title{Deep Clustering With Intra-class Distance Constraint for Hyperspectral Images}

\author{
	Jinguang~Sun, Wanli~Wang, Xian~Wei,~\IEEEmembership{Member,~IEEE}, Li~Fang,~\IEEEmembership{Member,~IEEE},\\
	Xiaoliang~Tang, Yusheng~Xu,~\IEEEmembership{Member,~IEEE}, Hui~Yu and Wei~Yao%
	\thanks{This work was partially supported by the Young Scientists Fund of the National Natural Science Foundation of China under Grant No.\ 61602226 and No.\ 61806186, and the CAS Pioneer Hundred Talents Program (Type C) under Grant No.\ 2017-122. \textit{(Jinguang~Sun and Wanli~Wang contribute equally to this work.) (Corresponding author: Xian~Wei.)}}%
	\thanks{J.~Sun and W.~Wang are with the School of Electronic and Information Engineering, Liaoning Technical University, Huludao 125105, Liaoning, China (email: Sunjinguang@lntu.edu.cn; email: wwlswj@163.com).}%
	\thanks{X.~Wei, L.~Fang, X.~Tang and H.~Yu are with the Quanzhou Institute of Equipment Manufacturing, Haixi Institute, Chinese Academy of Sciences, Quanzhou 362216, Fujian, China (email: xian.wei@fjirsm.ac.cn).}%
	\thanks{Y.~Xu is with the Department of Photogrammetry and Remote Sensing, Technical University of Munich, 80333 Munich, Germany.}%
	\thanks{W.~Yao is with the Department of Land Surveying and Geo-Informatics, Hong Kong Polytechnic University, 181 Chatham Road South, Hung Hom, Kowloon, Hong Kong.}
}

\markboth{}{Sun \MakeLowercase{\textit{et al.}}: Deep Clustering with Intra-class Distance Constraint for hyperspectral Images}

\maketitle

\begin{abstract}
	The high dimensionality of hyperspectral images often results in the degradation of clustering performance. Due to the powerful ability of deep feature extraction and non-linear feature representation, the clustering algorithm based on deep learning has become a hot research topic in the field of hyperspectral remote sensing. However, most deep clustering algorithms for hyperspectral images utilize deep neural networks as feature extractor without considering prior knowledge constraints that are suitable for clustering. To solve this problem, we propose an intra-class distance constrained deep clustering algorithm for high-dimensional hyperspectral images. The proposed algorithm constrains the feature mapping procedure of the auto-encoder network by intra-class distance so that raw images are transformed from the original high-dimensional space to the low-dimensional feature space that is more conducive to clustering. Furthermore, the related learning process is treated as a joint optimization problem of deep feature extraction and clustering. Experimental results demonstrate the intense competitiveness of the proposed algorithm in comparison with state-of-the-art clustering methods of hyperspectral images.
\end{abstract}


\IEEEpeerreviewmaketitle

\section{Introduction}

\IEEEPARstart{W}{ith} the development of the remote sensing technology, a wide diversity of sensor characteristics is nowadays available. The sensing data is ranging from medium and very high resolution (VHR) multispectral images to hyperspectral images that sample the electromagnetic spectrum with high detail \cite{li_performance_2016,romero_unsupervised_2016,yokoya2017hyperspectral,guo_enhanced_2017,zhai_kernel_2017,jiang_enhanced_2018,xie_unsupervised_2018,ghamisi2019multisource}. Utilizing these myriad sensors, the Earth Observation System (EOS) generates massive practical images of various land covering objects. Owing to abundant spatial and spectral information, these numerous images make it possible to extend the applications of hyperspectral remote sensing to many potential fields \cite{liu2016unsupervised,yu_generating_2016,zhang_deep_2016,zhao_spatial_2016,zhang_spectralspatial_2016,kale_research_2017,hong2019augmented}. However, it is very arduous to annotate these massive practical earth observation images effectively. Due to the lack of labeled high-dimensional samples of hyperspectral images, learning appropriate low-dimensional (LD) representations of data for clustering plays a critical role in hyperspectral image annotation and understanding.

Clustering is the task of grouping a set of objects in such a way that objects in the same group are more similar to each other than to those in other groups \cite{fahad_survey_2014}. For remote sensing images, the task is to classify the pixels into homogeneous regions which segment every image into different partitions \cite{alok_multi-objective_2016,li_performance_2016,xie_unsupervised_2018}. Most traditional clustering algorithms are based on shallow linear models \cite{patel_kernel_2014, yin_kernel_2016, zhang_self-weighted_2018, Wei2017Thesis}, such as algorithms based on K-means \cite{yang_two-stage_2017}, ISODATA \cite{hemalatha_unsupervised_2017} and Fuzzy C-means \cite{guo_enhanced_2017,jiang_enhanced_2018}, which often fail when the data exhibits an irregular non-linear distribution. During the past decades, spectral-based clustering methods \cite{shi_normalized_2000,ng_spectral_2002,von_luxburg_tutorial_2007,zhai_kernel_2017,zhai_new_2017, wei2016trace_cvpr} and density-based \cite{xie_unsupervised_2018} clustering methods have been state-of-the-arts. Let $\boldsymbol{X} = \left[\boldsymbol{x}_{1}, \cdots, \boldsymbol{x}_{n}\right] \in \mathbb{R}^{m \times n}$ be the matrix containing the $n$ independent training samples arranged as its columns, the spectral-based clustering approaches perform clustering in the following two steps. At first, an affinity matrix (i.e.,\ similarity graph) $\boldsymbol{C}$ is built to depict the relationship of the data, where $\boldsymbol{C}_{ij}$ denotes the similarity between data points $\boldsymbol{x}_i$ and $\boldsymbol{x}_j$. Secondly, the data is clustered through clustering the eigenvectors of the graph Laplacian $\boldsymbol{L}=\boldsymbol{D}^{-\frac{1}{2}}\boldsymbol{AD}^{-\frac{1}{2}}$, where $\boldsymbol{D}$ is a diagonal matrix with $\boldsymbol{D}_{ii}=\sum_{j}\boldsymbol{A}_{ij}$ and $\boldsymbol{A}=\left|\boldsymbol{C}\right|+\left|\boldsymbol{C}^T\right|$. The main idea of the density-based clustering \cite{ji_efficient_2014} approaches are to find high-density regions that are separated by low-density regions. The density peaks clustering algorithm (DPCA) proposed by Alex Rodriguez \cite{rodriguez_clustering_2014} has brought the density-based clustering approaches to a new stage. The core idea of DPCA is that the centre of the cluster is surrounded by some points with low local density, and these points are far away from other points with high local density. The DPCA separates the clustering process of non-clustered centre points into a single process. Due to the selection of the cluster centre and the classification of the non-cluster points are separated, the clustering precision is increased.

To solve the clustering problem of more complex distributed data, the sparse subspace clustering (SSC) algorithm \cite{elhamifar_sparse_2013, zhang_spectralspatial_2016}  is developed. The core idea of SSC is that among the infinite number of representations of a data point in terms of other points, a sparse representation corresponds to selecting a few points from the same subspace. In fact, the SSC algorithm is a solution of sparse optimization in the framework of spectral clustering, and it evaluates the labels for every sample in the low-dimensional space. Sharing the homogeneous idea with SSC, many sparse representation and low-rank approximation based methods for subspace clustering \cite{lee_acquiring_2005, patel_latent_2013, patel_kernel_2014, zhang_robust_2014, wei_trace_2018} have received a lot of attention in recent years. The key components of these methods are finding a sparse and low-rank representation of the data and then building a similarity graph on the sparse coefficient matrix for the separation of the data.

Although the spectral-based clustering algorithms and the density-based clustering algorithms can effectively cluster arbitrarily distributed data, only the shallow features of the data can be used \cite{yin_kernel_2016}. Moreover, it is difficult to further improve the clustering effect and precision. On the other hand, deep neural networks can non-linearly map data from the original feature space to a new feature space, with the promising advantages compared with traditional clustering algorithms for deep feature extraction and dimension reduction. Therefore, in recent years, the subspace clustering algorithm based on deep learning has attracted more attention.

The core idea of deep neural network clustering \cite{dundar_convolutional_2015,xie_unsupervised_2016,chang_deep_2017,ji_deep_2017,law_deep_2017} is to non-linearly map data from the original feature space to a new feature space and then complete clustering in the new feature space. Because its mapping is non-linear, deep neural network clustering has powerful capabilities on intrinsic feature extraction and data representation. The clustering algorithm based on auto-encoder networks \cite{song_auto-encoder_2013, dilokthanakul_deep_2016, chang_deep_2017, wei2018reconst_tnnls} is a popular framework for deep clustering algorithms, which utilizes a symmetric network structure to encode and represent the data. Such an algorithm consists of two important steps. Firstly, it obtains the code space of the data by reducing the data dimension and clusters the data in the obtained code space. Secondly, it performs the representation transformation from the obtained code space to a new generative feature space. Depending upon the auto-encoder network, the idea of generative adversarial \cite{goodfellow_generative_2014, hershey_deep_2016, liang_sub-gan:_2018, zhou_deep_2018, wei_adversarial_2018} is introduced into the clustering field, which further improves the performance of deep clustering algorithms.

Generally speaking, current deep clustering algorithms have the form \cite{lee_acquiring_2005, dilokthanakul_deep_2016, min_survey_2018, lin_deep_2018} as
\begin{equation}
\label{deep-clustering-form}
\mathcal{J}_{p}=\mathcal{J}_{p1} + \underbrace{\lambda\left\|\boldsymbol{Z}^{\left(\frac{M}{2}\right)} - \boldsymbol{Z}^{\left(\frac{M}{2}\right)}\boldsymbol{C}\right\|^2}_{\mathcal{J}_{p2}}+\mathcal{J}_{p3},
\end{equation}
where $\mathcal{J}_{p}$ is the loss function, $\mathcal{J}_{p1}$ represents the reconstruction error, $\mathcal{J}_{p2}$ is a sparse or low-rank constraint determined by the pre-trained matrix $\boldsymbol{C}$, and $\mathcal{J}_{p3}$ is the regularization term. The $\lambda$ in $\mathcal{J}_{p2}$ represents the constraint coefficient, and $\boldsymbol{Z}^{\left(\frac{M}{2}\right)}$ denotes the obtained codes or extracted features by the auto-encoder. The data representation shown in \eqref{deep-clustering-form} achieves both data reconstruction and dimensionality reduction. Therefore, the features extracted by such algorithms are appropriate ones for data dimensionality reduction. However, due to the lack of prior knowledge constraints, the feature extraction procedure of deep clustering algorithms tends to lose some useful guides which need to be further explored. To solve this problem, we propose an embedded deep clustering algorithm with intra-class distance constraint. The proposed algorithm embeds the global K-means model into the auto-encoder network that can constrain the procedure of data mapping and obtain data representation that is more conducive to clustering. The proposed algorithm has the following contributions:
\begin{enumerate}
    \item The intra-class distance is utilized to constrain the encoding process of the auto-encoder network so that the auto-encoder network can map the data from the original feature space to the feature space which is more conducive to clustering.
    \item The pre-training process is not necessary, and the indicator matrix is dynamically adjusted to image data.
    \item This work treats the solution to the proposed algorithm as a joint learning optimization problem, and the entire clustering procedure is completed in one stage.
\end{enumerate}

\section{Related Works}

In this section, we briefly discuss some existing works in unsupervised deep learning and subspace clustering.

\subsection{Auto-encoder Network}

With impressive learning and characterization capabilities, 
auto-encoder neural networks have achieved great success in various areas, especially in the scenario of unsupervised learning \cite{bengio_representation_2013, zhang_saliency-guided_2015, romero_unsupervised_2016, li_unsupervised_2017, wei2018reconst_tnnls}, such as natural language processing \cite{grozdic_whispered_2017}, image processing \cite{dai_analyzing_2018}, object detection \cite{park_multimodal_2018}, biometric recognition \cite{yu_multitask_2018}, and data analysis \cite{ma_deep_2018}. As state-of-the-art unsupervised techniques, auto-encoder and auto-encoder based neural networks also make outstanding contributions in the field of remote sensing. In this subsection, we briefly introduce the auto-encoder network.

\begin{figure}[!htbp]
    \centering
    \includegraphics[scale=0.32]{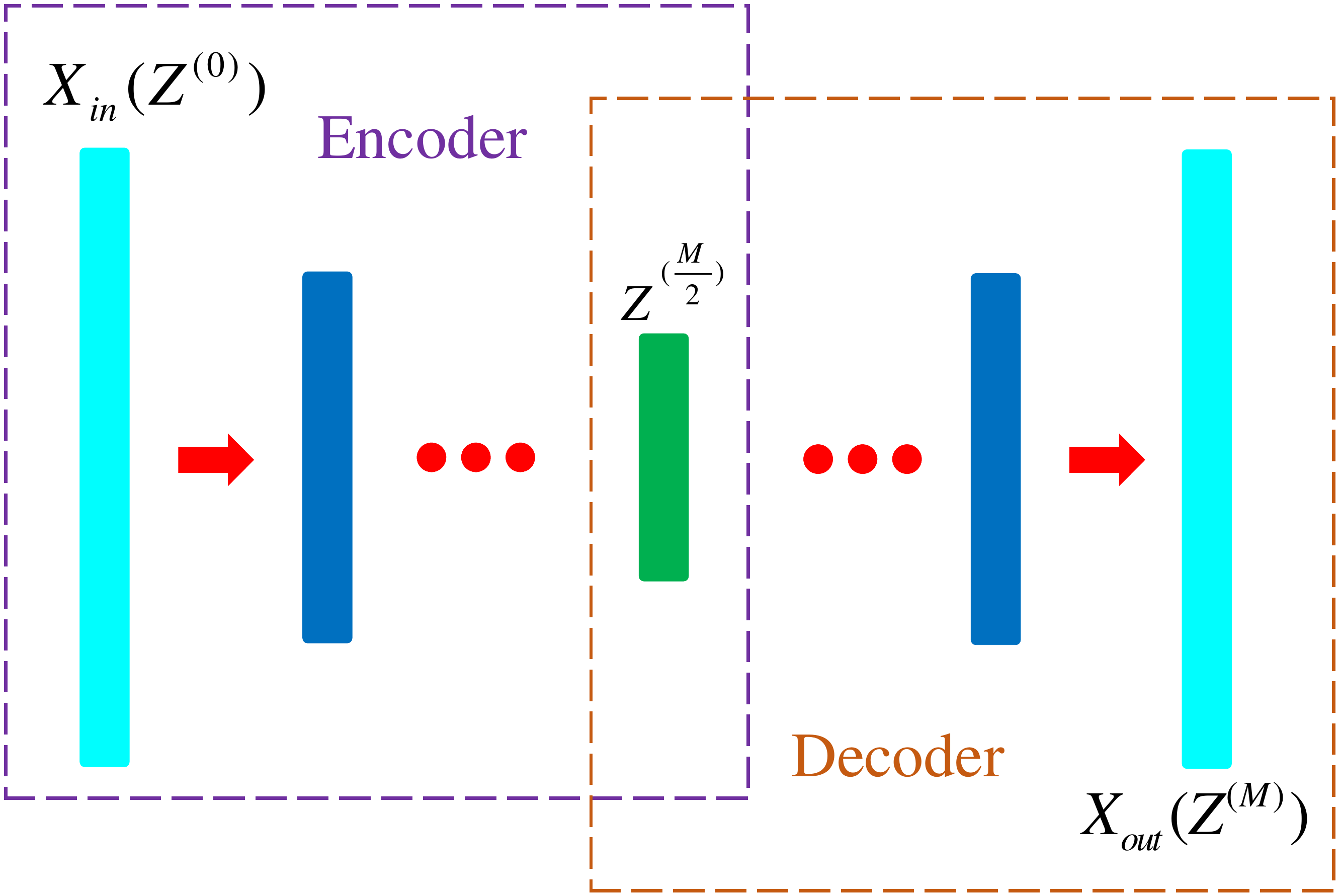}
    \caption{The structure of the auto-encoder network.}
    \label{basic-auto-encoder}
\end{figure}

In general, an auto-encoder \cite{song_auto-encoder_2013} is a kind of network that consists of encoder and decoder, and the structure of the encoder and decoder is symmetrical. If the auto-encoder contains multiple hidden layers, then the number of hidden layers of the encoder is equal to the number of hidden layers of the decoder. The structural model of the basic auto-encoder is shown in Fig.\ \ref{basic-auto-encoder}. The purpose of the basic auto-encoder is to reconstruct the input data at the output layer, and the perfect case is that the output signal $\boldsymbol{X}_{out}$ (i.e.,\ $\boldsymbol{Z}^{\left(M\right)}$) is exactly the same as the input signal $\boldsymbol{X}_{in}$ (i.e.,\ $\boldsymbol{Z}^{\left(0\right)}$). According to the structure shown in Fig.\ \ref{basic-auto-encoder}, the encoding process and decoding process of the basic auto-encoder can be described as
\begin{equation}\label{encoder-decoder}
\begin{cases}
\boldsymbol{z}^{\left(i+1\right)} = \mathcal{F}_e\left(\boldsymbol{W}_i\boldsymbol{z}^{\left(i\right)}+\boldsymbol{b}_i\right) \qquad &\text{Encoding}\\
\boldsymbol{z}^{\left(j+1\right)} = \mathcal{F}_d\left(\boldsymbol{W}_j\boldsymbol{z}^{\left(j\right)}+\boldsymbol{b}_j\right) \qquad &\text{Decoding}
\end{cases},
\end{equation}
where $\boldsymbol{W}_i$, $\boldsymbol{b}_i$ denote the $i$-$th$ encoding weight and the $i$-$th$ encoding bias, respectively, $\boldsymbol{W}_j$, $\boldsymbol{b}_j$ represent the $j$-$th$ decoding weight and the $j$-$th$ decoding bias, respectively, $\boldsymbol{z}^{\left( i\right)}$ denotes the data vector in $i$-$th$ layer, and $\mathcal{F}_e$ is the non-linear transformation. \emph{Sigmoid}, \emph{Tanh}, \emph{Relu} are commonly used activation functions for $\mathcal{F}_e$. $\mathcal{F}_d$ can be the same non-linear transformation as in the encoding process. Therefore, the loss function of the basic auto-encoder is to minimize the error between $\boldsymbol{X}_{in}$ and $\boldsymbol{X}_{out}$. The encoder converts the input signal into codes through some non-linear mapping, and the decoder tries to remap the codes to the input signal. The parameters of the auto-encoder, i.e.,\ weights and biases, are learned by minimizing the total reconstruction error, which could be computed by the mean square error
\begin{equation}\label{mean-square-error}
\begin{split}
\mathcal{J}_m\left(\boldsymbol{W},\boldsymbol{b}\right)&=\sum_{i=1}^{n}\mathcal{L}\left(\boldsymbol{z}_i^{\left(0\right)},\boldsymbol{z}_i^{\left(M\right)} \right)\\
&=\sum_{i=1}^{n}\left\| \boldsymbol{z}_i^{\left(0\right)} - \boldsymbol{z}_i^{\left(M\right)}\right\|_2^2, 
\end{split}
\end{equation}
or the cross entropy
\begin{equation}\label{cross-entropy}
\resizebox{0.85\hsize}{!}{$\begin{split}
    \mathcal{J}_c\left(\boldsymbol{W},\boldsymbol{b}\right)=&\sum_{i=1}^{n}\mathcal{L}\left(\boldsymbol{z}_i^{\left(0\right)},\boldsymbol{z}_i^{\left(M\right)}\right)
    \\=&-\sum_{i=1}^{n}\left(\boldsymbol{z}_i^{\left(0\right)}log\boldsymbol{z}_i^{\left(M\right)} + 
    \left(\boldsymbol{1}-\boldsymbol{z}_i^{\left(0\right)}\right)log\left(\boldsymbol{1}-\boldsymbol{z}_i^{\left(M\right)} \right)\right).
    \end{split}$}
\end{equation}

For the structure shown in Fig.\ \ref{basic-auto-encoder}, the hidden layers of the basic automatic coding network have three different structures: a compressed structure, a sparse structure, and an equivalent-dimensional structure. When the number of input layer neurons is greater than the number of hidden layer neurons, it is called a compressed structure \cite{peng_deep_2016}. Conversely, when the number of input layer neurons is smaller than the number of hidden layer neurons, it is called a sparse structure. If the input layer and the hidden layer have the same number of neurons, it is named the equivalent-dimensional structure.

\subsection{Deep Subspace Clustering}

Many practical subspace clustering missions transform data from the high dimensional feature space to the low dimensional feature space. Whereas many subspace clustering algorithms \cite{elhamifar_sparse_2013,liu_robust_2013,patel_kernel_2014,peng_robust_2015,yin_kernel_2016, wei_trace_2018} are shallow models, which cannot model the non-linearity of data in practical situations. Benefiting from the powerful capabilities of non-linear modeling
and data representation of the deep neural network, some clustering approaches based on deep neural networks have been proposed in recent years. Song et al.\ \cite{song_auto-encoder_2013} integrated an auto-encoder network with K-means to cluster the latent features. However, the feature mapping and clustering are two relatively independent processes in their work, and the K-means algorithm is not jointed into the feature mapping process. Therefore, the feature mapping process may not be constrained by the K-means algorithm. With the emergence of generative adversarial networks, some deep clustering algorithms \cite{xie_unsupervised_2016,zhou_deep_2018} embedded discriminant and adversarial ideas have been proposed, which further enhance the ability of deep feature extraction and data representation. On the other hand, some deep subspace clustering algorithms with prior constraints have been developed. According to different constraining conditions, these prior constraints may be sparse \cite{peng_deep_2016}, low rank \cite{chen_subspace_2018}, and least square \cite{ji_deep_2017}. Based on the auto-encoder network, the loss functions of these algorithms share the same modality, shown in \eqref{deep-subspace-clustering-function}. These algorithms learn representation for the input data with minimal reconstruction error and incorporate prior information into the potential feature learning to preserve the key reconstruction relation over the data. 
Although the features extracted by such algorithms are appropriate for the dimensionality reduction and reconstruction of the data, the procedure of feature extracting may lose the purposefulness, due to the lack of the constraints of prior knowledge.
\begin{equation}
\label{deep-subspace-clustering-function}
\begin{split}
\mathcal{J}_p\left(\boldsymbol{W}_i,\boldsymbol{b}_i\right) &=\underbrace{\frac{1}{2}\left\|\boldsymbol{Z}^{\left(0\right)}-\boldsymbol{Z}^{\left(M\right)}\right\|_F^2}_{\mathcal{J}_{p1}} \\&+\underbrace{\frac{\lambda_1}{2}\left\|\boldsymbol{Z}^{\left(\frac{M}{2}\right)}-\boldsymbol{Z}^{\left(\frac{M}{2}\right)}\boldsymbol{C}\right\|_F^2}_{\mathcal{J}_{p2}} \\&+\underbrace{\frac{\lambda_2}{2}\sum_{i=1}^{M}\left(\left\|\boldsymbol{W}_i\right\|_F^2+\left\|\boldsymbol{b}_i\right\|_2^2\right)}_{\mathcal{J}_{p3}},
\end{split}
\end{equation}
where $\mathcal{J}_{p1}$ denotes the reconstruction error, $\mathcal{J}_{p2}$ denotes the prior constraint, $\mathcal{J}_{p3}$ 
denotes the regularization term and $\boldsymbol{C}$ is a pre-trained matrix. Aforementioned deep clustering algorithms have following three weaknesses: 
\begin{enumerate}
    \item Lacking prior constraints related to clustering task.
    \item The matrix $\boldsymbol{C}$ needs to be pre-trained, which may not be optimal for various data to be clustered.
    \item Once given, the matrix $\boldsymbol{C}$ is fixed, which cannot be optimized jointly with the network.
\end{enumerate}

Different from these existing works, we propose an approach that embeds intra-class distance into an auto-encoder network. The Indicative matrix and the parameters of the network are adaptively optimized simultaneously. The proposed approach utilizes intra-class distance to constrain the feature mapping process of the auto-encoder so that the deep features extracted from the source space are more conducive for clustering. More details of the proposed approach are described in the following sections.

\section{The Proposed Deep Clustering Approach}

In this section, we elaborate on the details of the proposed \emph{Deep Clustering with Intra-class Distance Constraint} (DCIDC) algorithm. The framework of DCIDC is an auto-encoder network, and the specific structure may be varied according to different scenarios. With the constraint of intra-class distance, DCIDC extracts deep features of the data by mapping them from the source space to a latent feature space. In the new latent feature space, objects in the same group are more similar to each other than to those in other groups. 
We first explain how DCIDC is specifically designed and then present the algorithm for optimizing the DCIDC model.

\subsection{Deep Clustering with Intra-class Distance Constraint}

\begin{figure}[!htbp]
	\centering
	\includegraphics[scale=0.28]{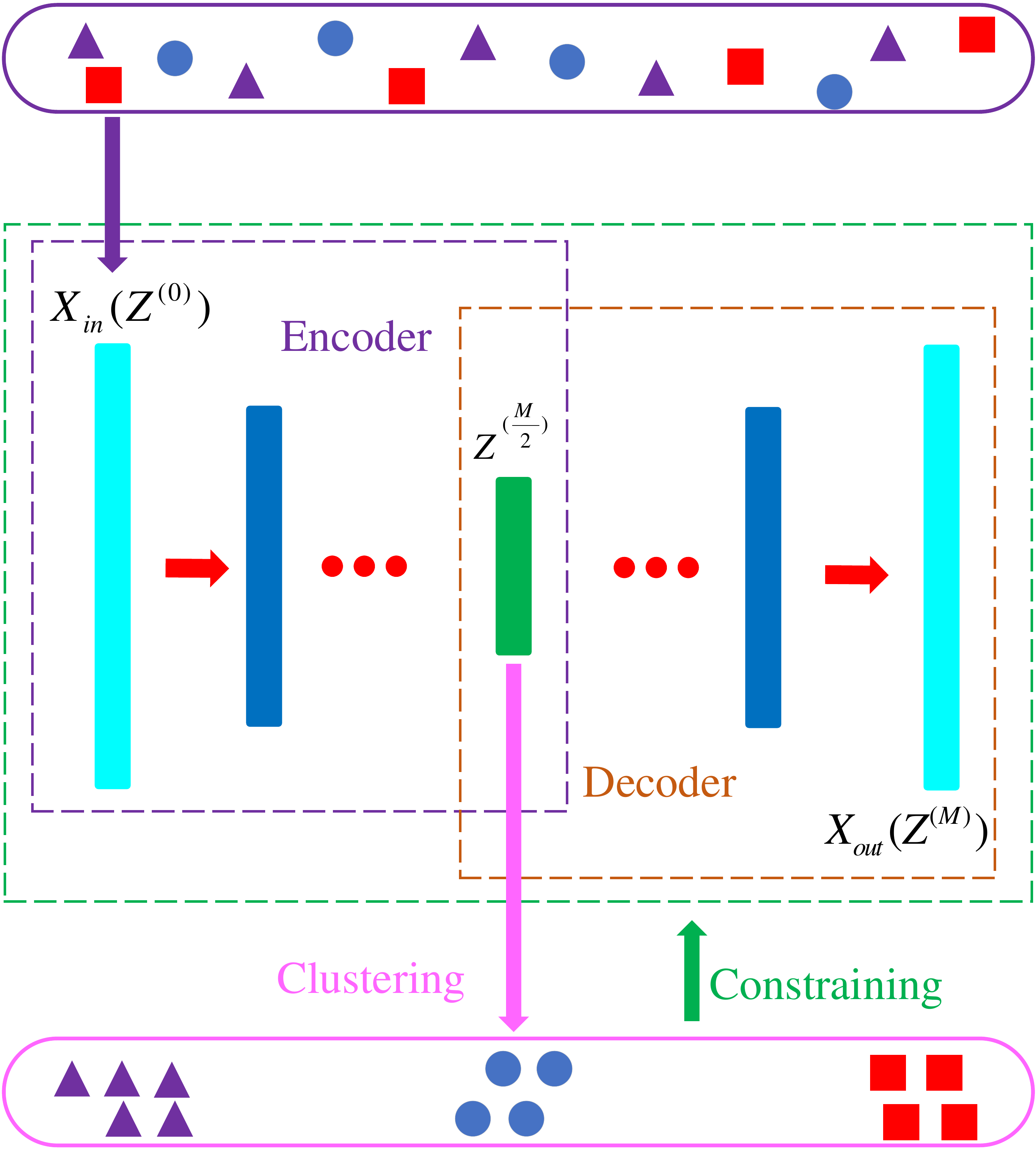}
	\caption{Network framework of the proposed algorithm}
	\label{DCIDC}
\end{figure}

The neural network within DCIDC consists of $M+1$ layers for performing $M$ non-linear transformations, where $M$ is an even number, 
the first $\frac{M}{2}$ hidden layers are encoders to learn a set of compact representations (i.e.,\ low-dimensional representations) 
and the last $\frac{M}{2}$ layers are decoders to progressively reconstruct the input. The framework of DCIDC is shown as Fig.\ \ref{DCIDC}. Let $\boldsymbol{Z}^{\left(0\right)}=\boldsymbol{X}_{in} \in \mathbb{R}^{N \times D}$ be one input matrix to the first layer, which denotes a hyperspectral image consisting of $N$ image pixels (samples) and $\boldsymbol{z}^{\left(0\right)}$ be one row of the matrix, which denotes a sample in $D$-dimensional feature space. For the encoder, the output of the $i$-$th$ layer is computed by
\begin{equation}\label{DCIDC-encoder}
\boldsymbol{z}^{\left(i\right)}=\mathcal{F}_e\left(\boldsymbol{W}_i\boldsymbol{z}^{\left(i-1\right)}+\boldsymbol{b}_i\right)\in\mathbb{R}^{d_i},
\end{equation}
where $i=1,2,\cdots,\frac{M}{2}$ indexes the layers of the encoder, $\boldsymbol{W}_i$ denotes the weight matrix from the $\left(i-1\right)$-$th$ layer to the $i$-$th$ layer and $\boldsymbol{b}_i$ denotes the bias of the $i$-$th$ layer. $\mathbb{R}^{d_i}$ indicates that the $\boldsymbol{z}^{\left(i\right)}$ belongs to a $d_i$-dimensional feature space. The $\mathcal{F}_e\left(\cdot\right)$ is a non-linear activation function. The $\frac{M}{2}$-$th$ layer $\boldsymbol{z}^{\left(\frac{M}{2}\right)}\in\mathbb{R}^{d_{\frac{M}{2}}}$ is shared by the encoder and the decoder. For the purpose of reducing the dimensionality of the input data, the dimensions of the layers in the encoder are designed to be $D \geq d_{i-1} \geq d_i \geq d_{\frac{M}{2}}$. For the decoder, the output of the $j$-$th$ layer can be computed by
\begin{equation}\label{DCIDC-decoder}
\boldsymbol{z}^{\left(j\right)}=\mathcal{F}_d\left(\boldsymbol{W}_j\boldsymbol{z}^{\left(j-1\right)}+\boldsymbol{b}_j\right)\in\mathbb{R}^{d_j},
\end{equation}
where $j=\frac{M}{2}+1,\frac{M}{2}+2,\cdots,M$ indexes the layers of the decoder and the non-linear activation function $\mathcal{F}_d\left(\cdot\right)$ can be the same as $\mathcal{F}_e\left(\cdot\right)$ or another absolutely different non-linear model. For the purpose of data reconstruction, the dimensions of the layers in the decoder are designed to be $d_{\frac{M}{2}} \leq d_{j-1} \leq d_j \leq d_M=D$. Thus, given a sample $\boldsymbol{z}^{\left(0\right)}$ (i.e.,\ $\boldsymbol{x}_{in}$) as one input of the first layer of DCIDC, $\boldsymbol{z}^{\left(M\right)}$ (i.e.,\ $\boldsymbol{x}_{out}$) is the reconstruction of $\boldsymbol{z}^{\left(0\right)}$, and the corresponding $\boldsymbol{z}^{\left(\frac{M}{2}\right)}$ is the representation of $\boldsymbol{x}_{in}$. Furthermore, for a data matrix $\boldsymbol{Z}^{\left(0\right)}=\left[\boldsymbol{z}_1^{\left(0\right)},\boldsymbol{z}_2^{\left(0\right)},\cdots,\boldsymbol{z}_N^{\left(0\right)}\right]^T\in\mathbb{R}^{N\times D}$ which denotes a collection fo $N$ given samples, the output matrix of the decoder $\boldsymbol{Z}^{\left(M\right)}=\left[\boldsymbol{z}_1^{\left(M\right)},\boldsymbol{z}_2^{\left(M\right)},\cdots,\boldsymbol{z}_N^{\left(M\right)}\right]^T\in\mathbb{R}^{N\times D}$ is the corresponding reconstruction for $\boldsymbol{Z}^{\left(0\right)}$, and the $\boldsymbol{Z}^{\left(\frac{M}{2}\right)}=\left[\boldsymbol{z}_1^{\left(\frac{M}{2}\right)},\boldsymbol{z}_2^{\left(\frac{M}{2}\right)},\cdots,\boldsymbol{z}_N^{\left(\frac{M}{2}\right)}\right]^T\in\mathbb{R}^{N\times d_{\frac{M}{2}}}$ is the desired low-dimensional representation of $\boldsymbol{Z}^{\left(0\right)}$.

The objective of DCIDC is to minimize the data reconstruction error and jointly constrain the non-linear transformation from $\boldsymbol{X}_{in}$ to the corresponding representation $\boldsymbol{Z}^{\left(\frac{M}{2}\right)}$ by intra-class distance. Thus, these targets can be formally stated as
\begin{equation}\label{DCIDC-loss-function}
\begin{split}
\min_{\boldsymbol{W}_i,\boldsymbol{b}_i}\mathcal{J}\left(\boldsymbol{W}_i,\boldsymbol{b}_i\right)&=\underbrace{\left\|\boldsymbol{Z}^{\left(0\right)}-\boldsymbol{Z}^{\left(M\right)}\right\|_F^2}_{\mathcal{J}_1}\\&+\underbrace{\lambda_1\left\|\boldsymbol{Z}^{\left(\frac{M}{2}\right)}-\boldsymbol{HS}^T\right\|_F^2}_{\mathcal{J}_2}\\&+\underbrace{\lambda_2\sum_{i=1}^{M}\left(\left\|\boldsymbol{W}_i\right\|_F^2+\left\|\boldsymbol{b}_i\right\|_2^2\right)}_{\mathcal{J}_3},
\end{split}
\end{equation}
where $\lambda_1$ and $\lambda_2$ are positive trade-off parameters. The terms $\mathcal{J}_1$, $\mathcal{J}_2$, and $\mathcal{J}_3$ are respectively designed for different goals. Intuitively, the first term $\mathcal{J}_1$ is designed for preserving locality by minimizing the reconstruction errors w.r.t.\ the input itself. In other words, the input acts as a supervisor for the procedure of learning a low-dimensional representation $\boldsymbol{Z}^{\left(\frac{M}{2}\right)}$. For the purpose that objects in the same cluster have similar features, the term $\mathcal{J}_2$ is designed to constrain the non-linear transformation from $\boldsymbol{Z}^{\left(0\right)}$ to its corresponding representation $\boldsymbol{Z}^{\left(\frac{M}{2}\right)}$, by minimizing the clustering error in each iteration. The matrix $\boldsymbol{S} \in \mathbb{R}^{d_{\frac{M}{2}}\times K}$ in \eqref{S} denotes the clustering centres, of which each column represents one cluster centre. The matrix $\boldsymbol{H} \in \mathbb{R}^{N\times K}$ in \eqref{H} is the indicative matrix and each row demonstrates the binary label. In other words, for each row of $\boldsymbol{H}$, there is only one element is $1$ and the rest are $0$. The $\boldsymbol{H}$ is not fixed and it is updated in each iteration. The $K$ in \eqref{S} and \eqref{H} denote the number of clusters. At last, $\mathcal{J}_3$ is a regularization term to avoid over-fitting.
\begin{equation}\label{S}
\boldsymbol{S}=
\begin{bmatrix}
s_{11}&\cdots&s_{1K}\\
\vdots&\ddots&\vdots\\
s_{d_{\frac{M}{2}}1}&\cdots&s_{d_{\frac{M}{2}}K}
\end{bmatrix}.
\end{equation}
\begin{equation}\label{H}
\boldsymbol{H}=
\begin{bmatrix}
0&1&0&\cdots&0\\
1&0&0&\cdots&0\\
\vdots&\vdots&\vdots&\ddots&\vdots\\
1&0&0&\cdots&0\\
0&0&0&\cdots&1
\end{bmatrix}.
\end{equation}

Our neural network model uses the input as self-supervisor to learn low-dimensional representations and jointly constrain the non-linear transformation by minimizing the clustering error, which is expected to enhance the deep intrinsic features extracted from the source data. The learned representations are fully adaptive and favorable for the clustering process. Furthermore, our model completes clustering in one step without additional pre-training process, which improves the efficiency.

\subsection{Optimization Procedure}

In this subsection, we mainly demonstrate how the proposed DCIDC model can be optimized efficiently via gradient descent and the solution procedure of $\boldsymbol{H}$. As the optimization of parameters $\boldsymbol{W}$ and $\boldsymbol{b}$ does not share the same mechanism with $\boldsymbol{H}$ and $\boldsymbol{S}$, we present the gradient descent and the calculation of $\boldsymbol{H}$ and $\boldsymbol{S}$ respectively. For the convenience of developing the algorithm, we rewrite \eqref{DCIDC-loss-function} in the following sample-wise form.
\begin{equation}\label{DCIDC-loss-function-sample}
\resizebox{0.85\hsize}{!}{$\begin{split}
	\mathcal{J}&=\frac{1}{2}\sum_{i=1}^{N}\left(\left\|\boldsymbol{z}_i^{\left(0\right)}-\boldsymbol{z}_i^{\left(M\right)}\right\|_2^2+\lambda_1\left\|\boldsymbol{z}_i^{\left(\frac{M}{2}\right)}-\boldsymbol{h}_i\boldsymbol{S}^T\right\|_2^2\right)\\&+\frac{\lambda_2}{2}\sum_{m=1}^{M}\left(\left\|\boldsymbol{W}_m\right\|_F^2+\left\|\boldsymbol{b}_m\right\|_2^2\right).
	\end{split}$}
\end{equation}
According to the definition of $\boldsymbol{z}^{\left(i\right)}$ in \eqref{DCIDC-encoder}, $\boldsymbol{z}^{\left(i\right)}$ in \eqref{DCIDC-decoder} and the chain rule, we can express the gradients of \eqref{DCIDC-loss-function-sample} w.r.t.\ $\boldsymbol{W}_m$ and $\boldsymbol{b}_m$ as \eqref{partial-w} and \eqref{partial-b}, respectively.
\begin{equation}\label{partial-w}
\frac{\partial\mathcal{J}}{\partial\boldsymbol{W}_m}=\left(\boldsymbol{\Delta}_m+\lambda_1\boldsymbol{\Lambda}_m\right)\left(\boldsymbol{z}_i^{\left(m-1\right)}\right)^T+\lambda_2\boldsymbol{W}_m,
\end{equation}
\begin{equation}\label{partial-b}
\frac{\partial\mathcal{J}}{\partial\boldsymbol{b}_m}=\boldsymbol{\Delta}_m+\lambda_1\boldsymbol{\Lambda}_m+\lambda_2\boldsymbol{b}_m,
\end{equation}
where $\boldsymbol{\Delta}_m$ is defined as
\begin{equation}\label{delta-definition}
\resizebox{0.85\hsize}{!}{$\boldsymbol{\Delta}_m=
	\begin{cases}
	-\left(\boldsymbol{z}_i^{\left(0\right)}-\boldsymbol{z}_i^{\left(M\right)}\right)\bigodot\mathcal{G}^{'}\left(\boldsymbol{y}_i^{\left(M\right)}\right)&\quad m=M\\
	\left(\boldsymbol{W}_{m+1}\right)^T\boldsymbol{\Delta}_{m+1}\bigodot\mathcal{G}^{'}\left(\boldsymbol{y}_i^{\left(m\right)}\right)&\quad\text{Otherwise}
	\end{cases},$}
\end{equation}
and $\boldsymbol{\Lambda}_m$ is given as
\begin{equation}\label{lambda-definition}
\resizebox{0.85\hsize}{!}{$\boldsymbol{\Lambda}_m=
	\begin{cases}
	\left(\boldsymbol{W}_{m+1}\right)^T\boldsymbol{\Lambda}_{m+1}\bigodot\mathcal{G}^{'}\left(\boldsymbol{y}_i^{\left(m\right)}\right)&\quad m=1,\cdots,\frac{M}{2}-1\\
	\left(\boldsymbol{z}_i^{\left(\frac{M}{2}\right)} - \boldsymbol{h}_i\boldsymbol{S}^T\right)\bigodot\mathcal{G}^{'}\left(\boldsymbol{y}_i^{\left(\frac{M}{2}\right)}\right)&\quad m=\frac{M}{2}\\
	\boldsymbol{0}&\quad m=\frac{M}{2}+1,\cdots,M
	\end{cases},$}
\end{equation}
where the $\bigodot$ denotes element-wise multiplication, $\boldsymbol{y}_i^{\left(m\right)}=\boldsymbol{W}_m\boldsymbol{z}_i^{\left(m-1\right)}+\boldsymbol{b}_m$, and $\mathcal{G}^{'}\left(\cdot\right)$ is the derivative of the activation function $\mathcal{G}\left(\cdot\right)$ defined as
\begin{equation}\label{active-function-g}
\mathcal{G}\left(\cdot\right)=
\begin{cases}
\mathcal{F}_e\left(\cdot\right)\qquad&m=1,\cdots,\frac{M}{2}\\
\mathcal{F}_d\left(\cdot\right)\qquad&m=\frac{M}{2}+1,\cdots,M
\end{cases}.
\end{equation}

Using the gradient descent algorithm, we update $\{\boldsymbol{W}_m,\boldsymbol{b}_m\}_{m=1}^M$ as \eqref{W-update} and \eqref{b-update} until convergence.
\begin{equation}\label{W-update}
\boldsymbol{W}_m\leftarrow\boldsymbol{W}_m-\mu\frac{\partial\mathcal{J}}{\partial\boldsymbol{W}_m},
\end{equation}
\begin{equation}\label{b-update}
\boldsymbol{b}_m\leftarrow\boldsymbol{b}_m-\mu\frac{\partial\mathcal{J}}{\partial\boldsymbol{b}_m},
\end{equation}
where $\mu>0$ is the learning rate which is typically set to a small value according to specific scenarios.

As the output of DCIDC model, the indicative matrix $\boldsymbol{H}$ is calculated via the least distance rule in each clustering step. In other words, $\boldsymbol{H}$ is updated by solving the term \eqref{equation-get-h} in every iteration.
\begin{equation}\label{equation-get-h}
\min_{\boldsymbol{H},\boldsymbol{S}}\left\|\boldsymbol{Z}^{\left(\frac{M}{2}\right)}-\boldsymbol{HS}^T\right\|_F^2.
\end{equation}
Thus, with the accomplishment of the optimization of DCIDC model, the final $\boldsymbol{H}$ will be simultaneously obtained. Now, we demonstrate the solution of \eqref{equation-get-h}.

For the task of clustering, the matrix $\boldsymbol{Z}^{\left(\frac{M}{2}\right)}$ can be redefined as
\begin{equation}\label{Z-M-2-redefined}
	\boldsymbol{Z}^{\left(\frac{M}{2}\right)}=\{\boldsymbol{Z}_i^{\left(\frac{M}{2}\right)}\}_{i=1}^K,
\end{equation}
where $\boldsymbol{Z}_i^{\left(\frac{M}{2}\right)}$ is a component of $\boldsymbol{Z}^{\left(\frac{M}{2}\right)}$, which denotes one cluster of the data, $\bigcup_{i=1}^K\boldsymbol{Z}_i^{\left(\frac{M}{2}\right)}=\boldsymbol{Z}^{\left(\frac{M}{2}\right)}$ and $\boldsymbol{Z}_i^{\left(\frac{M}{2}\right)}\bigcap\boldsymbol{Z}_j^{\left(\frac{M}{2}\right)}=\boldsymbol{\varPhi}$ w.r.t.\ $i \neq j \in \{1, 2, \cdots, K\}$. Then, for the convenience of solving \eqref{equation-get-h}, it can be transformed into
\begin{equation}\label{sse-s}
	\min_{\boldsymbol{H}, \boldsymbol{S}}\mathcal{E}_{\boldsymbol{S}}=\sum_{i=1}^{K}\sum_{\boldsymbol{z}^{\left(\frac{M}{2}\right)} \in \boldsymbol{Z}_i^{\left(\frac{M}{2}\right)}}\left\|\boldsymbol{s}_i^T-\boldsymbol{z}^{\left(\frac{M}{2}\right)}\right\|_2^2,
\end{equation}
where $\boldsymbol{s}_i$ is a column vector of $\boldsymbol{S}$, which denotes the centre of $\boldsymbol{Z}_i^{\left(\frac{M}{2}\right)}$. Let
\begin{equation}\label{partial-s}
	\begin{split}
	\frac{\partial\mathcal{E}_{\boldsymbol{S}}}{\partial\boldsymbol{S}}&=\frac{\partial}{\partial\boldsymbol{S}}\sum_{i=1}^{K}\sum_{\boldsymbol{z}^{\left(\frac{M}{2}\right)} \in \boldsymbol{Z}_i^{\left(\frac{M}{2}\right)}}\left\|\boldsymbol{s}_i^T-\boldsymbol{z}^{\left(\frac{M}{2}\right)}\right\|_2^2\\
	&=\sum_{i=1}^{K}\sum_{\boldsymbol{z}^{\left(\frac{M}{2}\right)} \in \boldsymbol{Z}_i^{\left(\frac{M}{2}\right)}}\frac{\partial}{\partial\boldsymbol{S}}\left\|\boldsymbol{s}_i^T-\boldsymbol{z}^{\left(\frac{M}{2}\right)}\right\|_2^2=0,
	\end{split}
\end{equation}
we get
\begin{equation}\label{get-s}
	\boldsymbol{s}_i^T=\frac{1}{n_i}\sum_{\boldsymbol{z}^{\left(\frac{M}{2}\right)} \in \boldsymbol{Z}_i^{\left(\frac{M}{2}\right)}}\boldsymbol{z}^{\left(\frac{M}{2}\right)},
\end{equation}
where $n_i$ is the number of pixel of the cluster $\boldsymbol{Z}_i^{\left(\frac{M}{2}\right)}$.

Similarly, for the purpose of calculating $\boldsymbol{H}$, the \eqref{equation-get-h} can be rewritten as
\begin{equation}\label{sse-h}
	\min_{\boldsymbol{H}, \boldsymbol{S}}\mathcal{E}_{\boldsymbol{H}}=\sum_{i=1}^{K}\sum_{\boldsymbol{z}^{\left(\frac{M}{2}\right)} \in \boldsymbol{Z}_i^{\left(\frac{M}{2}\right)}}\left\|\left(\boldsymbol{z}^{\left(\frac{M}{2}\right)}\right)^T-\boldsymbol{S}\boldsymbol{h}^T\right\|_2^2,
\end{equation}
where $\boldsymbol{h}$ is the indicator corresponding to $\boldsymbol{z}^{\left(\frac{M}{2}\right)}$, which is serialized as a row vector of $\boldsymbol{H}$. Let
\begin{equation}\label{partial-h}
	\begin{split}
	&\frac{\partial}{\partial\boldsymbol{h}^T}\left\|\left(\boldsymbol{z}^{\left(\frac{M}{2}\right)}\right)^T-\boldsymbol{S}\boldsymbol{h}^T\right\|_2^2\\
	&=\frac{\partial}{\partial\boldsymbol{h}^T}\left\|\boldsymbol{S}\boldsymbol{h}^T-\left(\boldsymbol{z}^{\left(\frac{M}{2}\right)}\right)^T\right\|_2^2\\
	&=\frac{\partial}{\partial\boldsymbol{h}^T}[\boldsymbol{S}\boldsymbol{h}^T-\left(\boldsymbol{z}^{\left(\frac{M}{2}\right)}\right)^T]^T[\boldsymbol{S}\boldsymbol{h}^T-\left(\boldsymbol{z}^{\left(\frac{M}{2}\right)}\right)^T]\\
	&=2\boldsymbol{S}^T\boldsymbol{S}\boldsymbol{h}^T-2\boldsymbol{S}^T\left(\boldsymbol{z}^{\left(\frac{M}{2}\right)}\right)^T=0,
	\end{split}
\end{equation}
we get
\begin{equation}\label{get-h}
	\boldsymbol{h}^T=\left(\boldsymbol{S}^T\boldsymbol{S}\right)^{-1}\boldsymbol{S}^T\left(\boldsymbol{z}^{\left(\frac{M}{2}\right)}\right)^T.
\end{equation}
Thus, we get
\begin{equation}\label{get-H}
	\boldsymbol{H}=[\mathcal{T}\left(\boldsymbol{h}_1\right), \mathcal{T}\left(\boldsymbol{h}_2\right), \cdots, \mathcal{T}\left(\boldsymbol{h}_N\right)]^T,
\end{equation}
where
\begin{equation}\label{T-function}
	\mathcal{T}\left(\boldsymbol{h}\right)=\mathcal{T}\{h_i\}_{i=1}^K=
	\begin{cases}
	1 &\quad h_i=\text{\emph{Max}}\left(\{h_i\}_{i=1}^K\right)\\
	0 &\quad \text{Otherwise}
	\end{cases}.
\end{equation}
So far, the detailed procedure for optimizing the proposed model DCIDC can be summarized as Algorithm \ref{alg1}.
\begin{algorithm}
	\caption{Algorithm of Deep Clustering with Intra-class Distance Constraint}
	\label{alg1}
	\begin{algorithmic}[1]
		\REQUIRE The data matrix $\boldsymbol{X}_{in}$ (i.e.,\ $\boldsymbol{Z}^{\left(0\right)}$) and the number of cluster $K$.
		\ENSURE The indicative matrix $\boldsymbol{H}$.
		\STATE Initialize a matrix $\boldsymbol{H}$ according to \eqref{H} and the given $K$.
		\FOR{$i=1$ to $M$}
		\STATE Initialize $\boldsymbol{W}_i$.
		\STATE Initialize $\boldsymbol{b}_i$.
		\ENDFOR
		\WHILE{not convergence}
		\FOR{$i=1$ to $\frac{M}{2}$}
		\STATE $\boldsymbol{z}^{\left(i\right)}\leftarrow\mathcal{F}_e\left(\boldsymbol{W}_i\boldsymbol{z}^{\left(i-1\right)}+\boldsymbol{b}_i\right)$.
		\ENDFOR
		\FOR{$j=\frac{M}{2}+1$ to $M$}
		\STATE $\boldsymbol{z}^{\left(j\right)}\leftarrow\mathcal{F}_d\left(\boldsymbol{W}_j\boldsymbol{z}^{\left(j-1\right)}+\boldsymbol{b}_j\right)$.
		\ENDFOR
		\STATE Calculate $\mathcal{J}_1$ in \eqref{DCIDC-loss-function}.
		\STATE Calculate $\boldsymbol{s}_i$ by \eqref{get-s} and $\boldsymbol{S} \gets [\boldsymbol{s}_1, \boldsymbol{s}_2, \cdots, \boldsymbol{s}_K]$.
		\STATE Calculate $\mathcal{J}_2$ in \eqref{DCIDC-loss-function}.
		\STATE Calculate $\mathcal{J}_3$ in \eqref{DCIDC-loss-function}.
		\STATE Update $\boldsymbol{H}$ by \eqref{get-H}.
		\FOR{$i=1$ to $M$}
		\STATE Update $\boldsymbol{W}_i$ by \eqref{W-update}.
		\STATE Update $\boldsymbol{b}_i$ by \eqref{b-update}.
		\ENDFOR
		\ENDWHILE
		\RETURN $\boldsymbol{H}$.
	\end{algorithmic}
\end{algorithm}

\section{Experimental Results}

In this section, we compare the proposed DCIDC approach with popular clustering methods on four image datasets in terms of two evaluation metrics and discuss the influence on DCIDC with different coefficient values of $\lambda_1$ and activation functions.

\subsection{Experimental Settings}
\subsubsection{Datasets}
We carry out our experiments using four hyperspectral image datasets: Indian Pines, Pavia, Salinas, and Salinas-A. The Indian Pines dataset was gathered by 224-band AVIRIS sensor over the Indian Pines test site in North-western Indiana. It consists of $145 \times 145$ pixels and 224 spectral reflectance bands in the wavelength range of $0.4 \sim 2.5 \times 10^{-6}$ meters. The number of bands of the Indian Pines dataset used in our experiment is reduced to 200 by removing bands covering the region of water absorption. The ROSIS sensor acquired the Pavia dataset during a flight campaign over Pavia, northern Italy. The used Pavia dataset in our experiment is a $1096 \times 1096$ pixels image with 100 spectral bands. The AVIRIS sensor collected the Salinas dataset over Salinas Valley, California, characterized by the high spatial resolution. The area covered comprises 512 lines by 217 samples. As with Indian Pines scene, 24 water absorption bands are discarded. A small sub-scene of Salinas image, denoted as Salinas-A, is adopted too.

\subsubsection{Evaluation Criteria}
We adopt two metrics to evaluate the clustering quality: accuracy and normalized mutual information(NMI). Higher values of these metrics indicate better performance. For each dataset, we repeat each algorithm five times and report the means and the standard deviations of these metrics.

\subsubsection{Baseline Algorithms}
For the sake of fairness, we compare DCIDC with clustering algorithms that carried out experiments with the same four datasets: Indian Pines, Pavia, Salinas, and Salinas-A. These algorithms are the deep subspace clustering with sparsity Prior (PARTY), the auto-encoder based subspace clustering (AESSC), the sparse subspace clustering (SSC), the latent subspace sparse subspace clustering (LS3C), the low-rank representation based clustering (LRR), the low rank based subspace clustering (LRSC), and the smooth representation clustering (SMR). Among these methods, PARTY and AESSC are deep clustering methods.

In the experiments with datasets Indian Pines, Salinas, and Salinas-A, the proposed DCIDC is designed as a seven layer neural network structure which consists of $200 \times 128 \times 64 \times 32 \times 64 \times 128 \times 200$ neurons. For the experiment with dataset Pavia, the network structure contains $100 \times 72 \times 36 \times 25 \times 36 \times 72 \times 100$ neurons. For fair comparisons, we report the best results of all the evaluated methods achieved with their optimal parameters. For our DCIDC with the trade-off parameters $\lambda_1$ and $\lambda_2$, we fix $\lambda_2=0.0003$ for all the data sets and experimentally choose $\lambda_1$.

\subsection{Comparison With The Evaluated Methods}
\begin{figure}[!htbp]
	\centering
	\subfloat[]
	{
		\centering
		\includegraphics[scale=0.3]{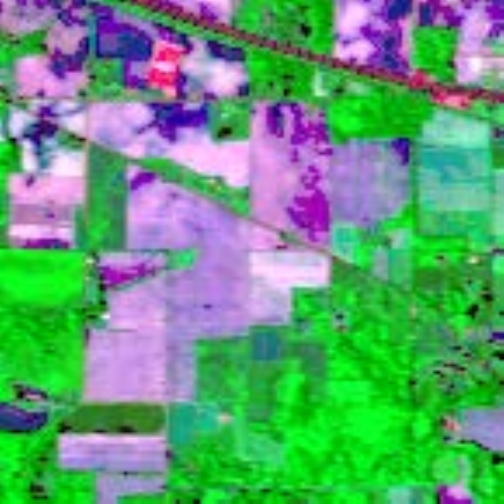}
	}
	\subfloat[]
	{
		\centering
		\includegraphics[scale=0.3]{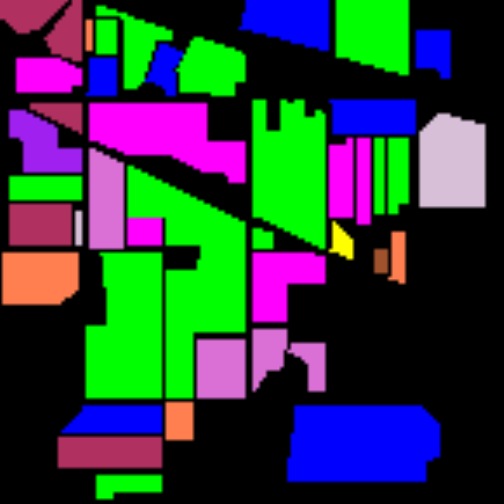}
	}
	\subfloat[]
	{
		\centering
		\includegraphics[scale=0.3]{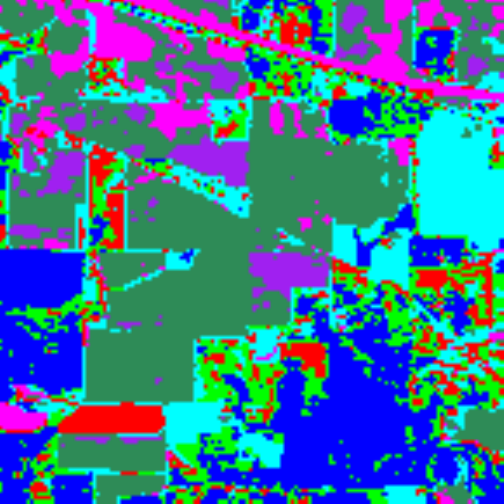}
	}
	\subfloat[]
	{
		\centering
		\includegraphics[scale=0.3]{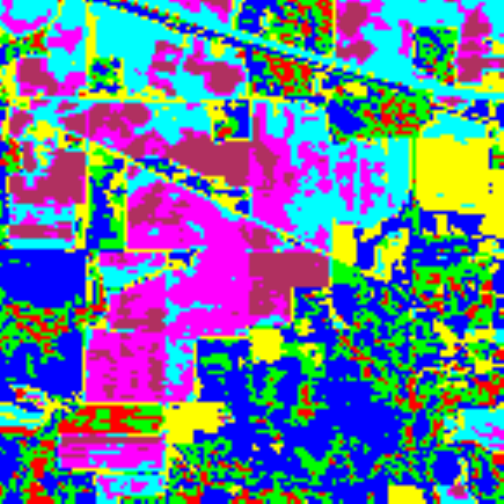}
	}
	\subfloat[]
	{
		\centering
		\includegraphics[scale=0.3]{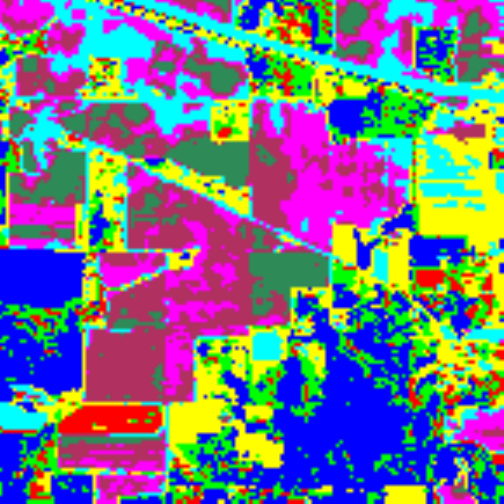}
	}\\
	\subfloat[]
	{
		\centering
		\includegraphics[scale=0.3]{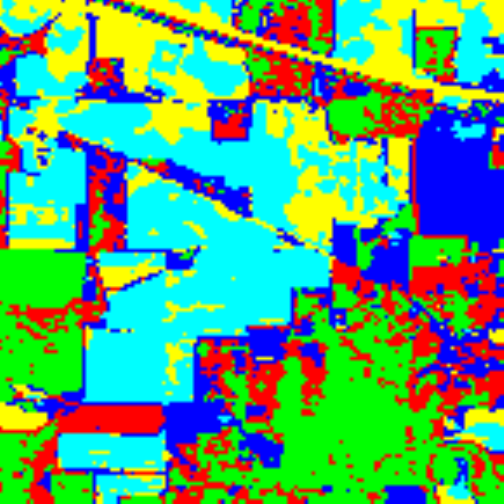}
	}
	\subfloat[]
	{
		\centering
		\includegraphics[scale=0.3]{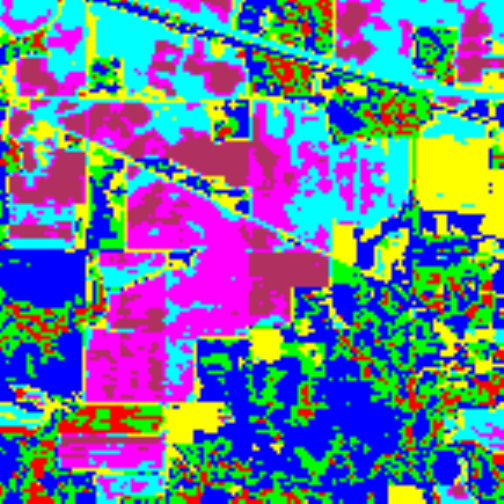}
	}
	\subfloat[]
	{
		\centering
		\includegraphics[scale=0.3]{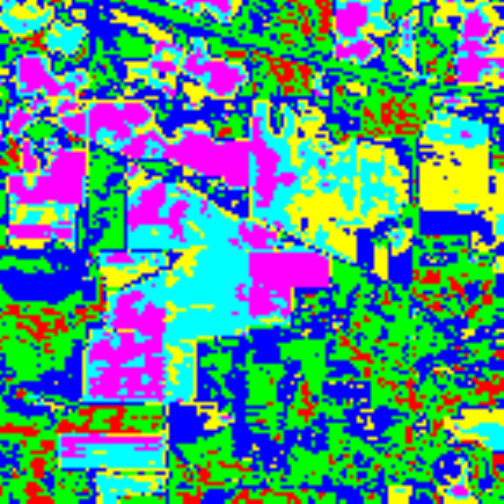}
	}
	\subfloat[]
	{
		\centering
		\includegraphics[scale=0.3]{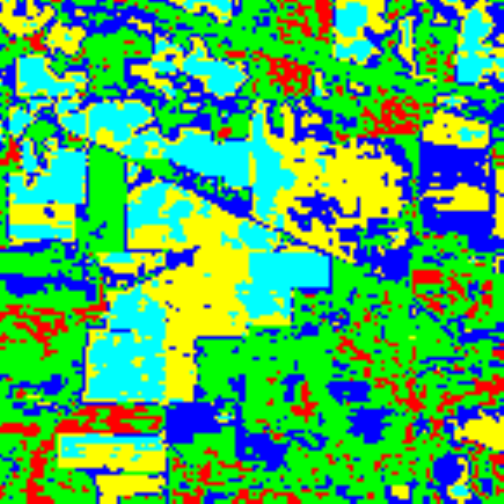}
	}
	\subfloat[]
	{
		\centering
		\includegraphics[scale=0.3]{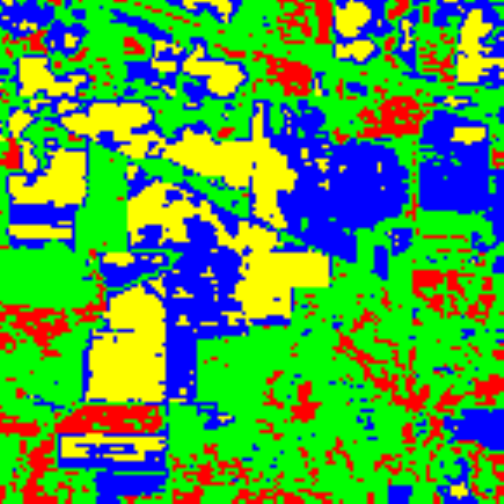}
	}\\
	\caption{The performance comparison of difierent algorithms on the dataset Indian Pines. (a) The Indian Pines image, (b) ground truth, (c) DCIDC, (d) PARTY, (e) AESSC, (f) SSC, (g) LS3C, (h) LRR, (i) LRSC, (j) SMR.}\label{result-inidan-pines}
\end{figure}

\begin{figure}[!htbp]
	\centering
	\subfloat[]
	{
		\centering
		\includegraphics[scale=0.2]{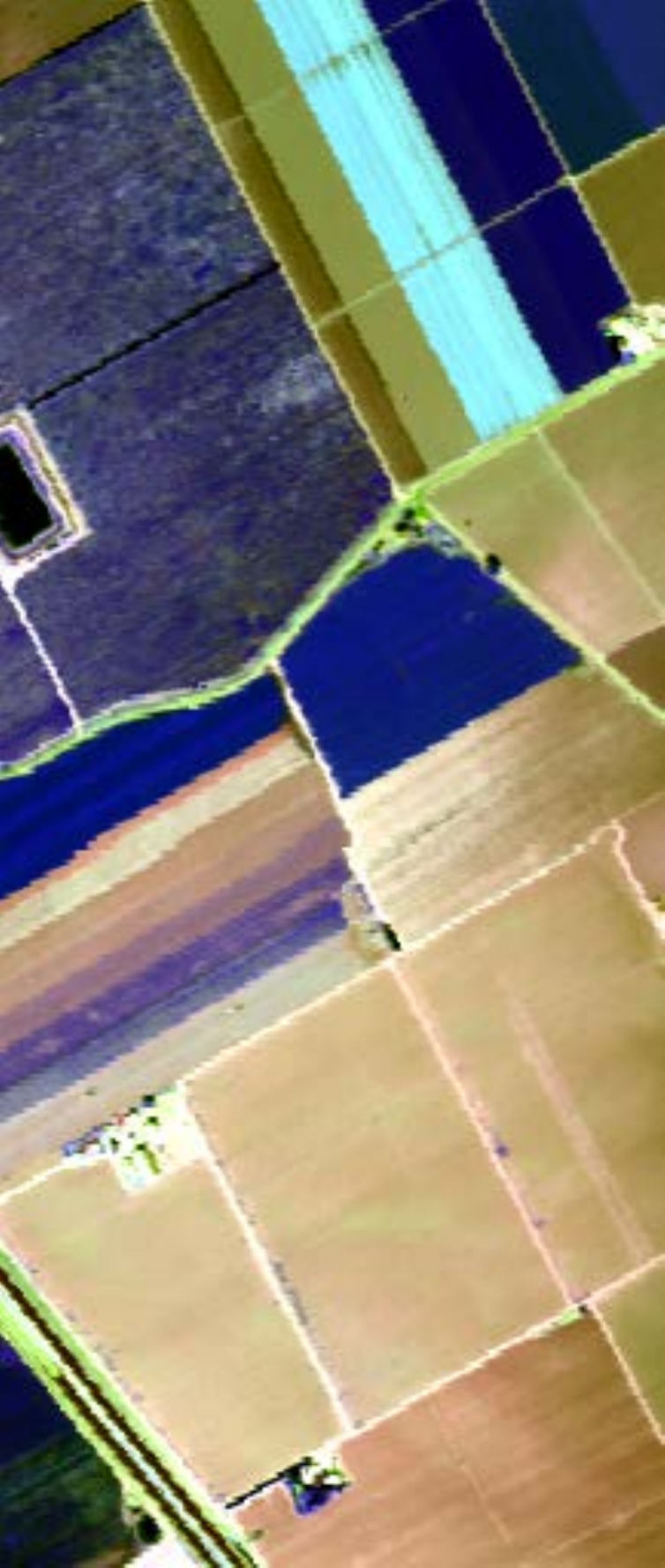}
	}
	\subfloat[]
	{
		\centering
		\includegraphics[scale=0.2]{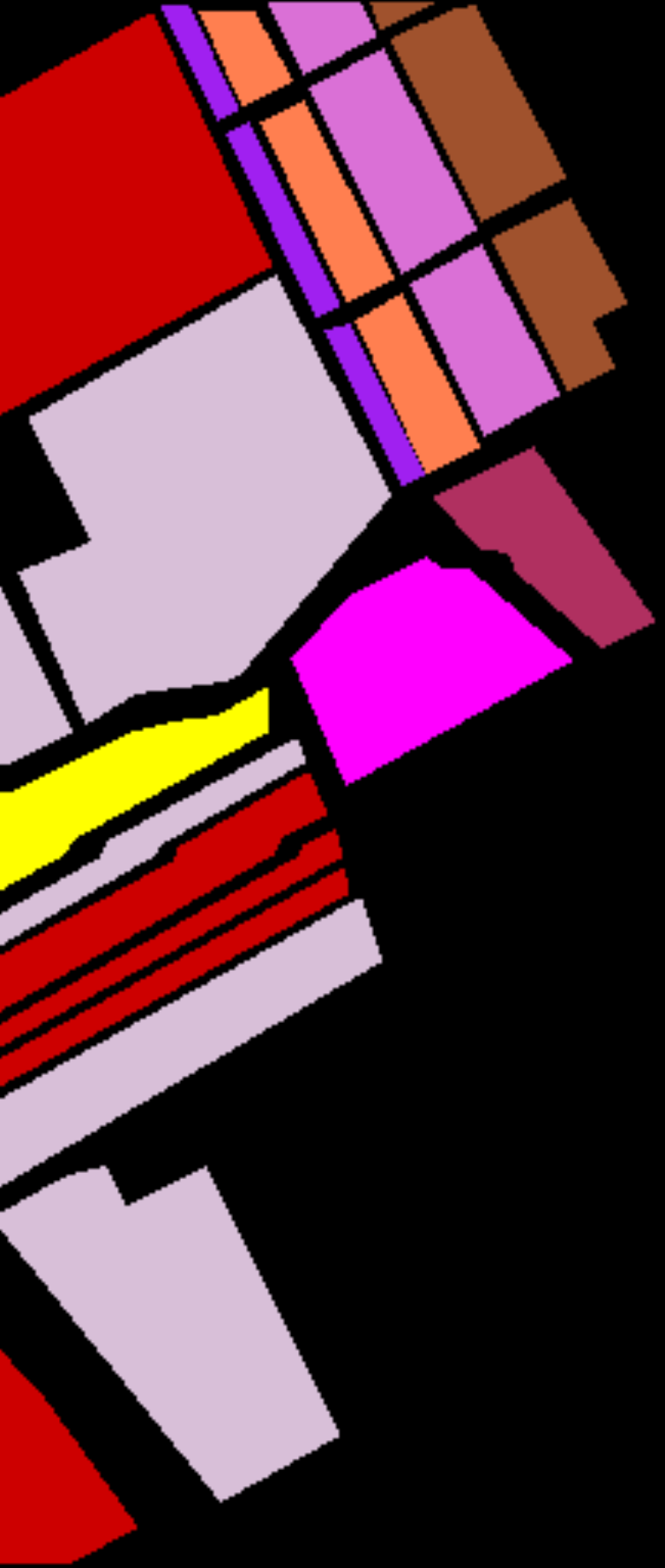}
	}
	\subfloat[]
	{
		\centering
		\includegraphics[scale=0.2]{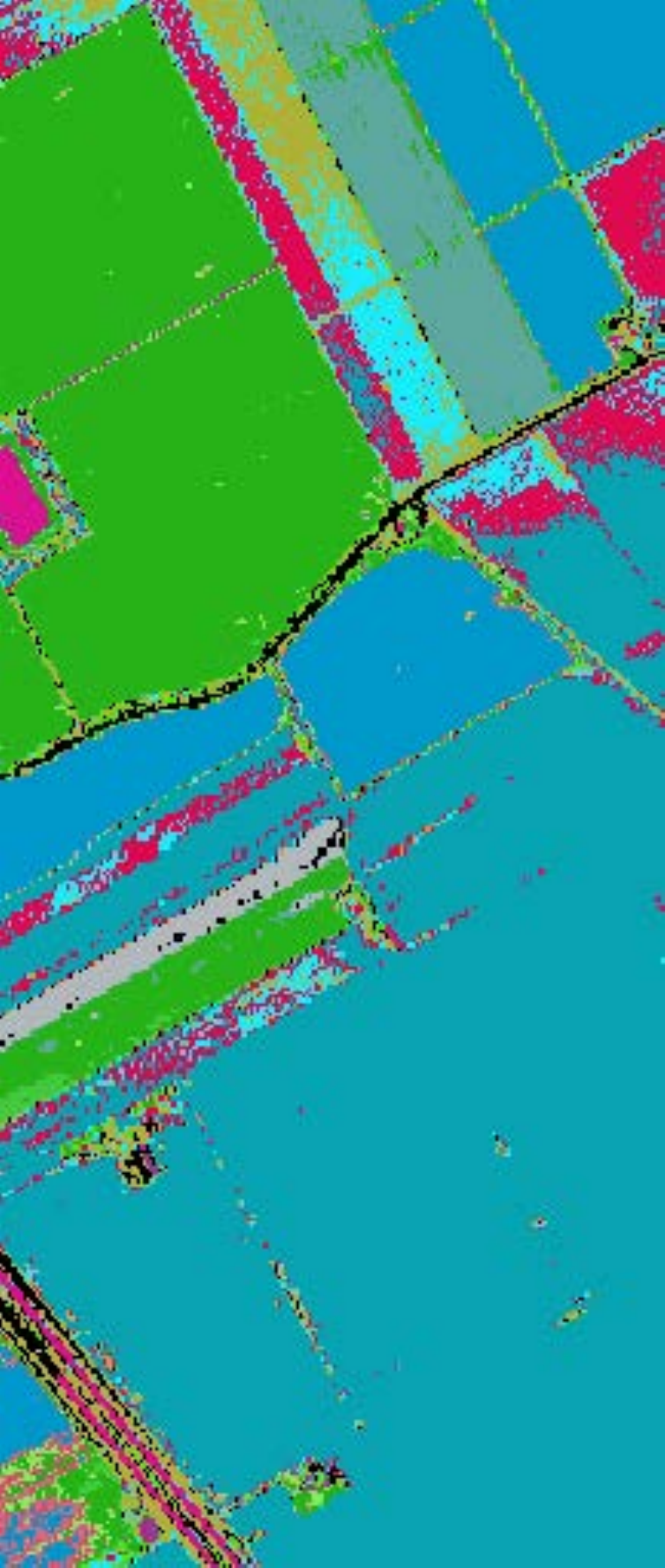}
	}
	\subfloat[]
	{
		\centering
		\includegraphics[scale=0.2]{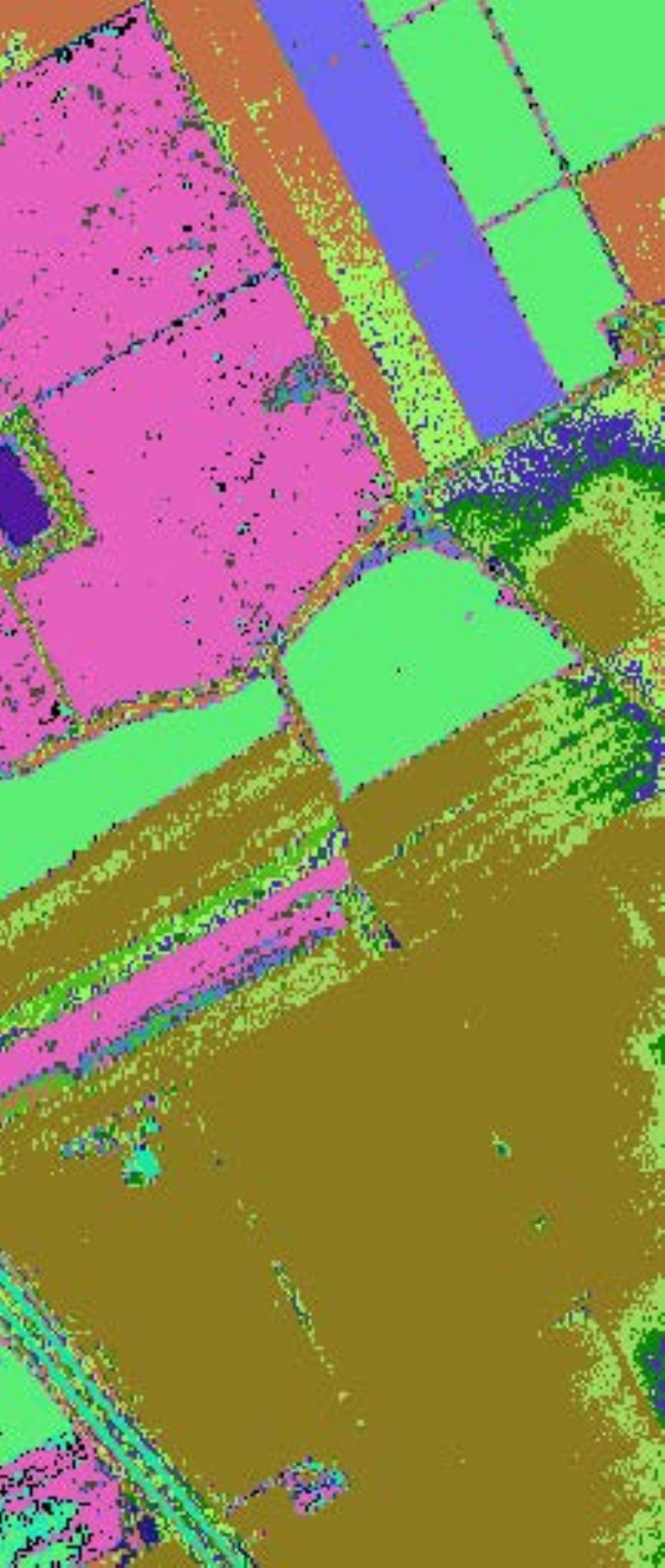}
	}
	\subfloat[]
	{
		\centering
		\includegraphics[scale=0.2]{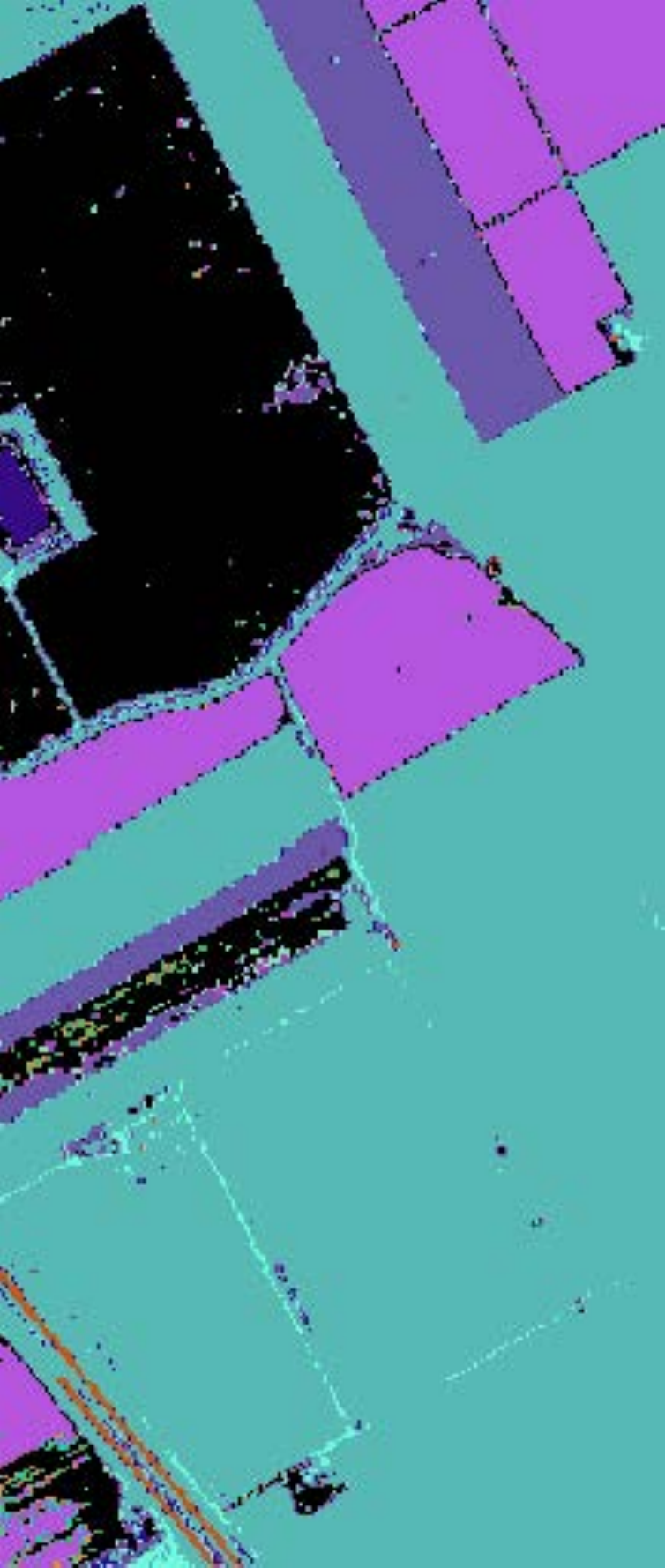}
	}\\
	\subfloat[]
	{
		\centering
		\includegraphics[scale=0.2]{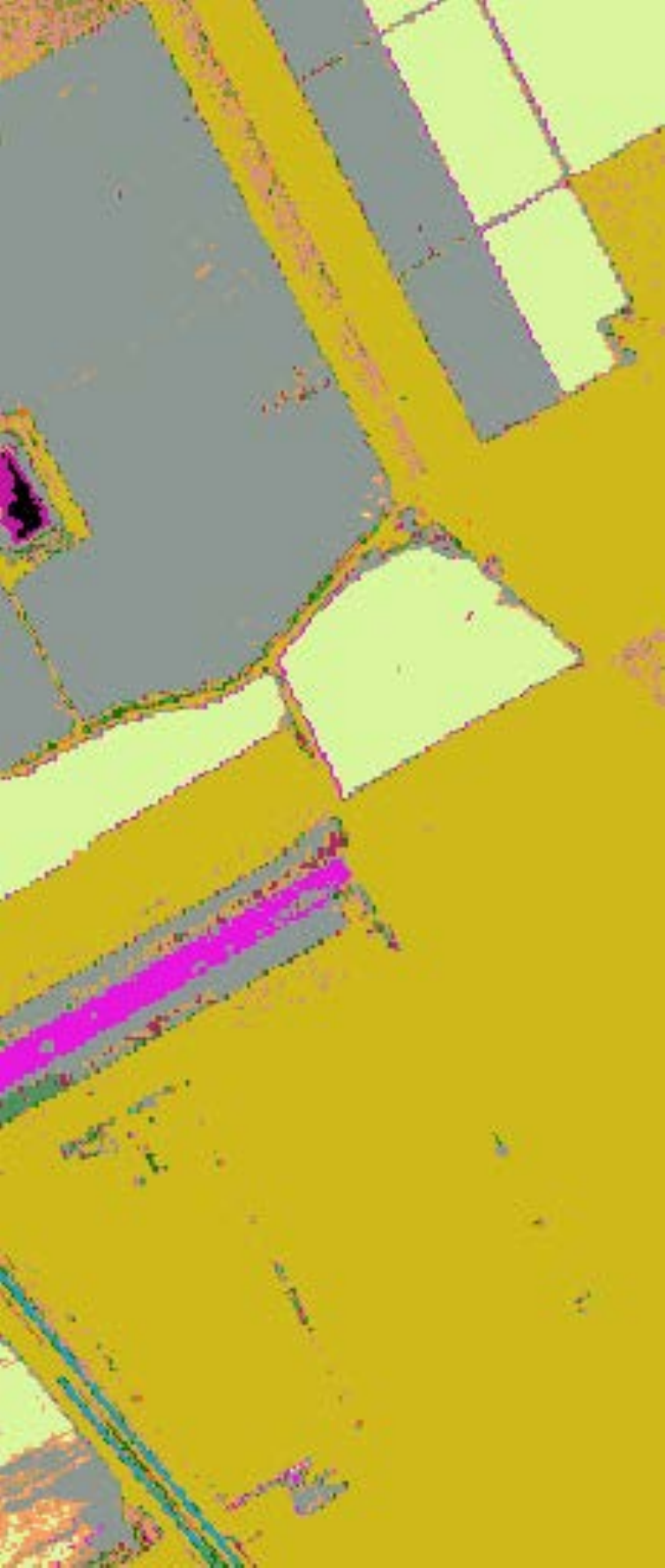}
	}
	\subfloat[]
	{
		\centering
		\includegraphics[scale=0.2]{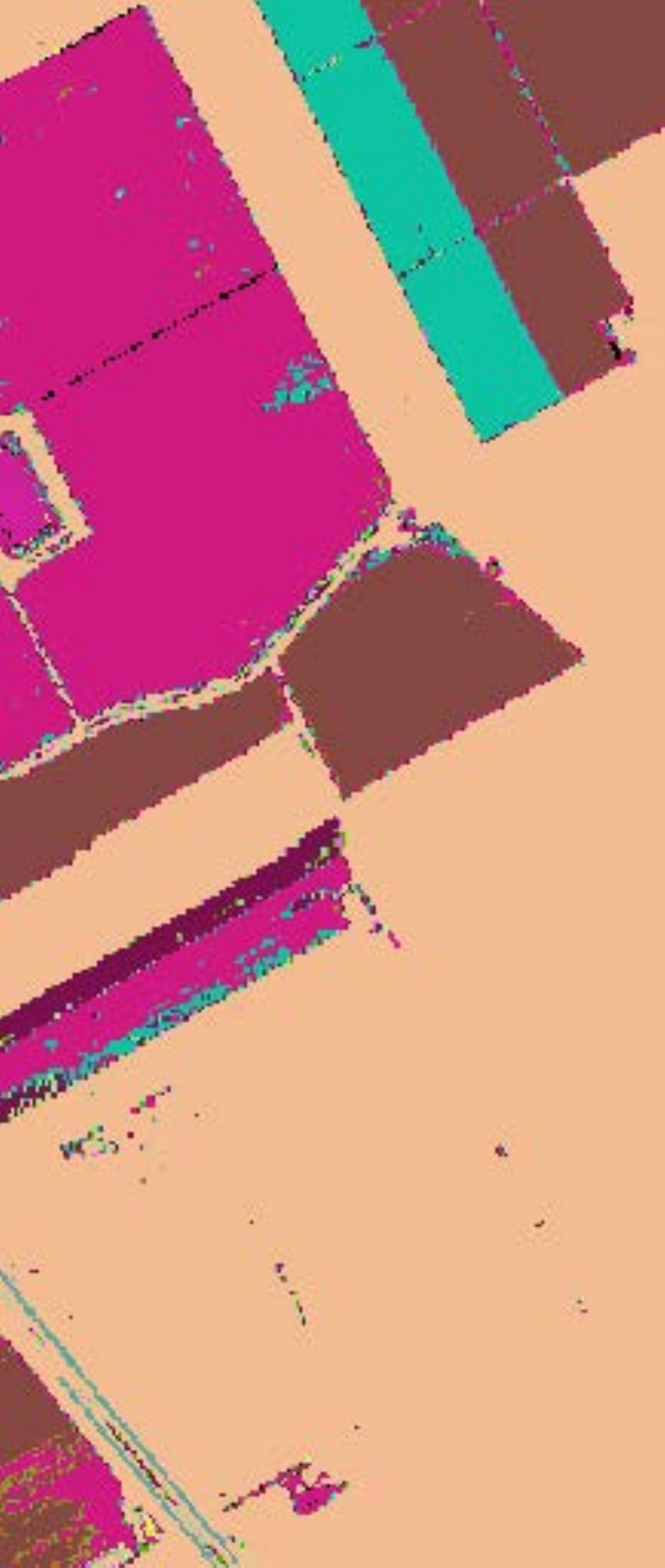}
	}
	\subfloat[]
	{
		\centering
		\includegraphics[scale=0.2]{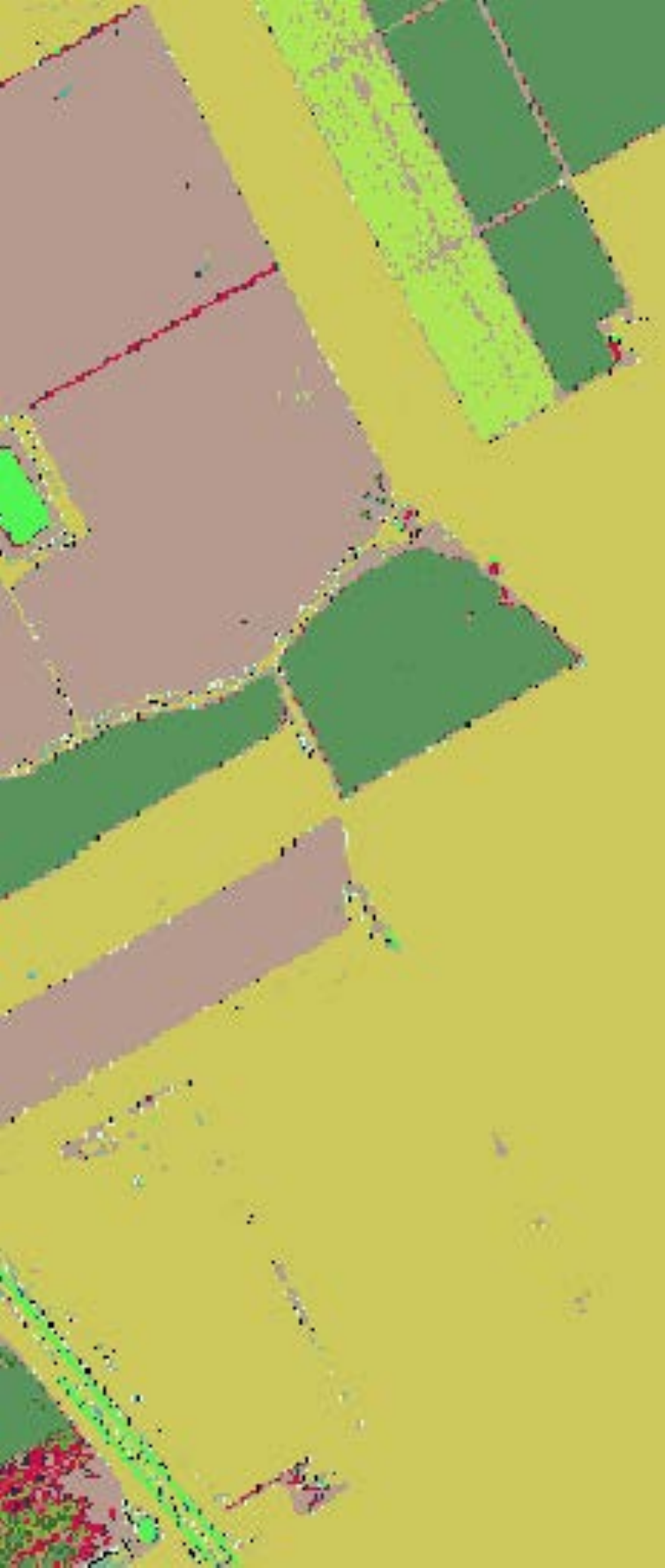}
	}
	\subfloat[]
	{
		\centering
		\includegraphics[scale=0.2]{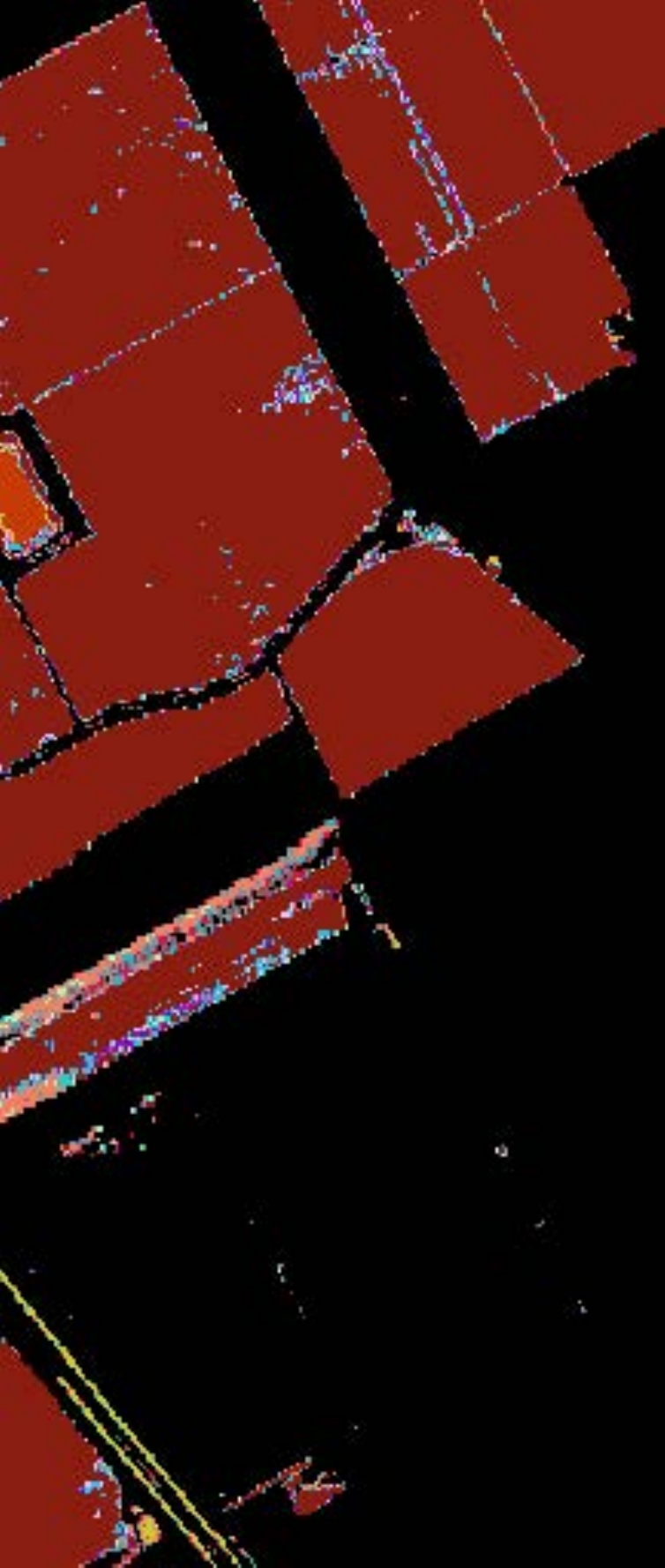}
	}
	\subfloat[]
	{
		\centering
		\includegraphics[scale=0.2]{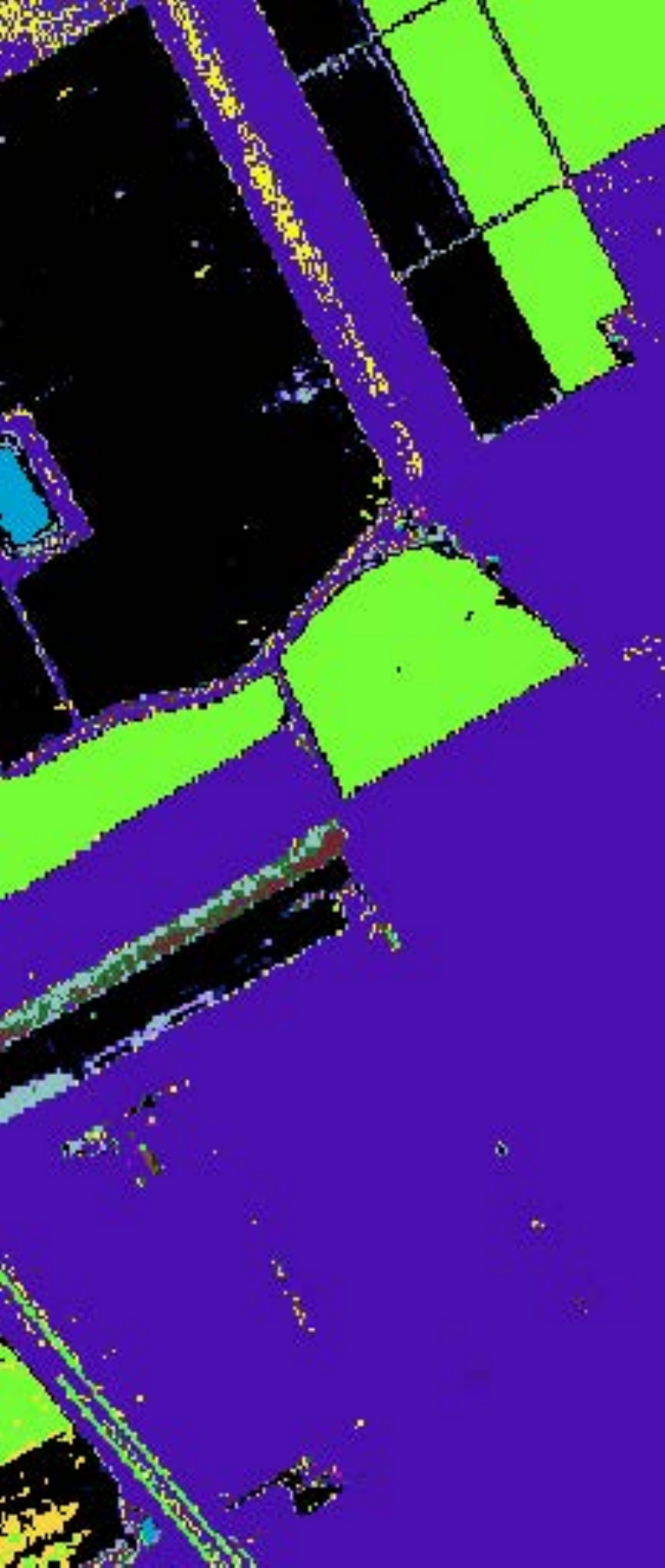}
	}\\
	\caption{The performance comparison of difierent algorithms on the dataset Salinas. (a) The Salinas image, (b) ground truth, (c) DCIDC, (d) PARTY, (e) AESSC, (f) SSC, (g) LS3C, (h) LRR, (i) LRSC, (j) SMR.}\label{result-salinas}
\end{figure}

\begin{figure}[!htbp]
	\centering
	\subfloat[]
	{
		\centering
		\includegraphics[scale=0.5]{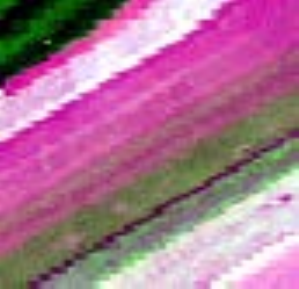}
	}
	\subfloat[]
	{
		\centering
		\includegraphics[scale=0.5]{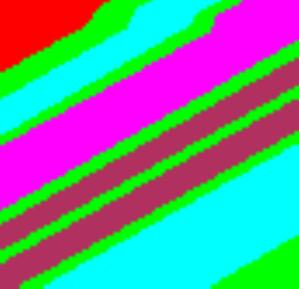}
	}
	\subfloat[]
	{
		\centering
		\includegraphics[scale=0.5]{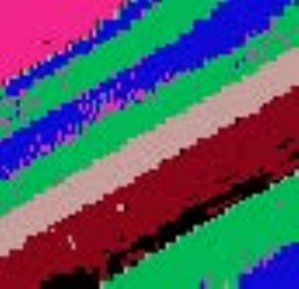}
	}
	\subfloat[]
	{
		\centering
		\includegraphics[scale=0.5]{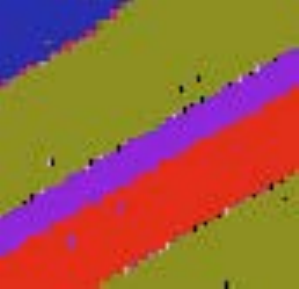}
	}
	\subfloat[]
	{
		\centering
		\includegraphics[scale=0.5]{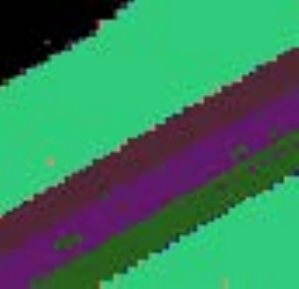}
	}\\
	\subfloat[]
	{
		\centering
		\includegraphics[scale=0.5]{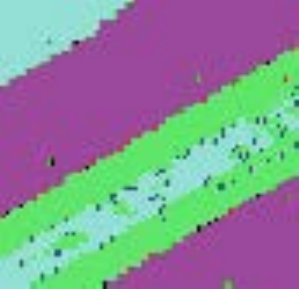}
	}
	\subfloat[]
	{
		\centering
		\includegraphics[scale=0.5]{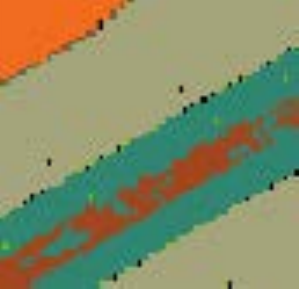}
	}
	\subfloat[]
	{
		\centering
		\includegraphics[scale=0.5]{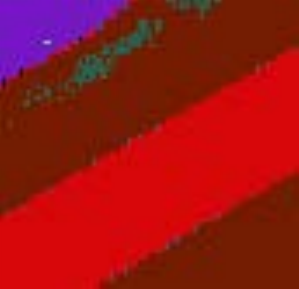}
	}
	\subfloat[]
	{
		\centering
		\includegraphics[scale=0.5]{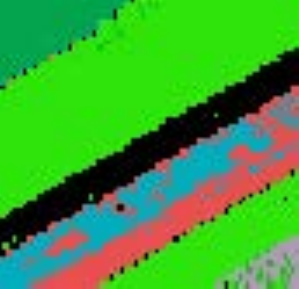}
	}
	\subfloat[]
	{
		\centering
		\includegraphics[scale=0.5]{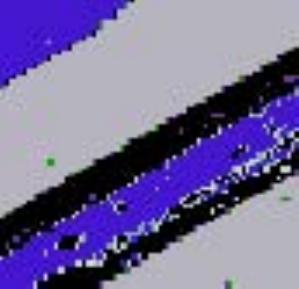}
	}\\
	\caption{The performance comparison of difierent algorithms on the dataset Salinas-A. (a) The Salinas-A image, (b) ground truth, (c) DCIDC, (d) PARTY, (e) AESSC, (f) SSC, (g) LS3C, (h) LRR, (i) LRSC, (j) SMR.}\label{result-salinasa}
\end{figure}

\begin{figure}[!htbp]
	\centering
	\subfloat[]
	{
		\centering
		\includegraphics[scale=0.0625]{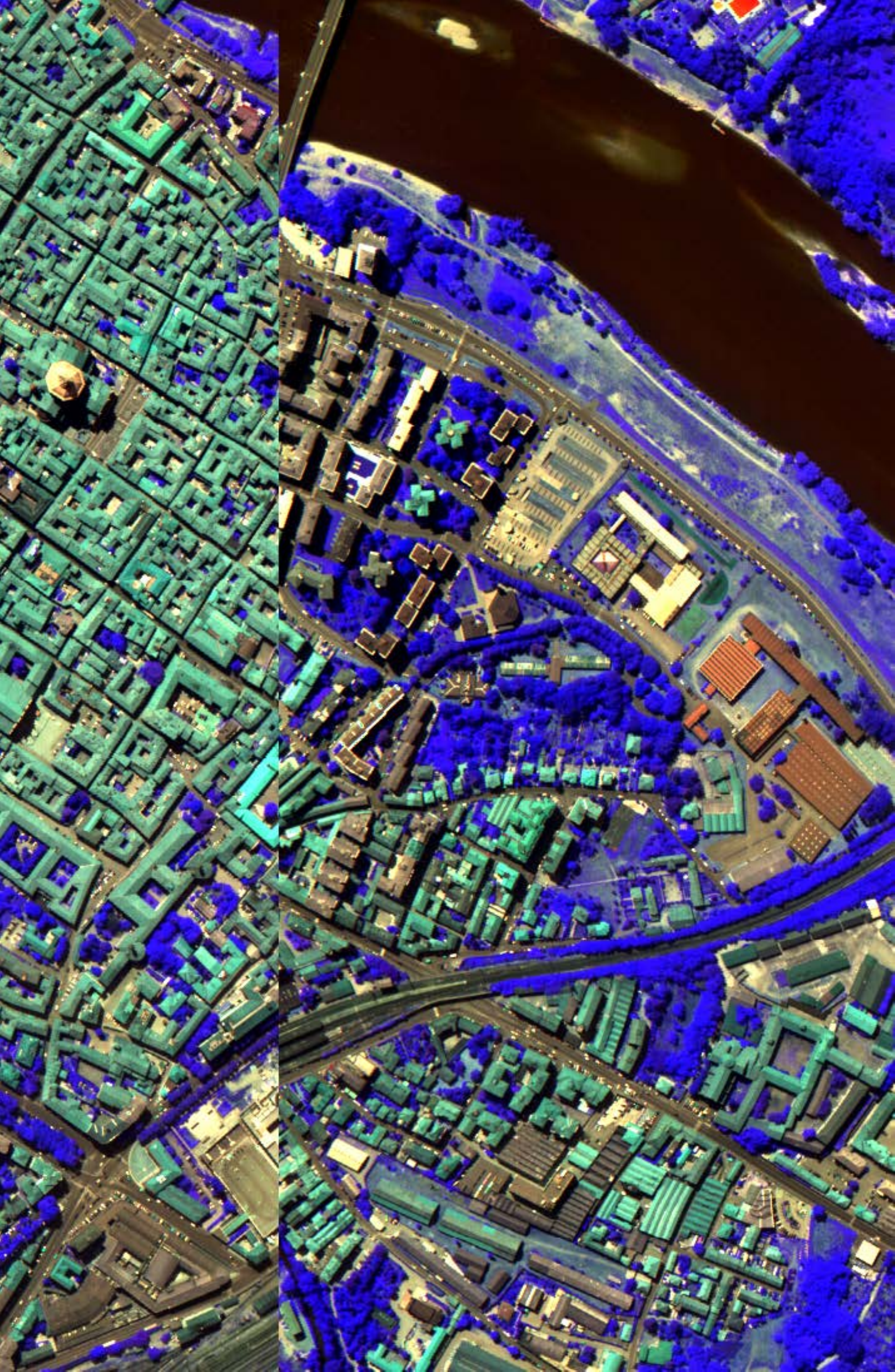}
	}
	\subfloat[]
	{
		\centering
		\includegraphics[scale=0.0625]{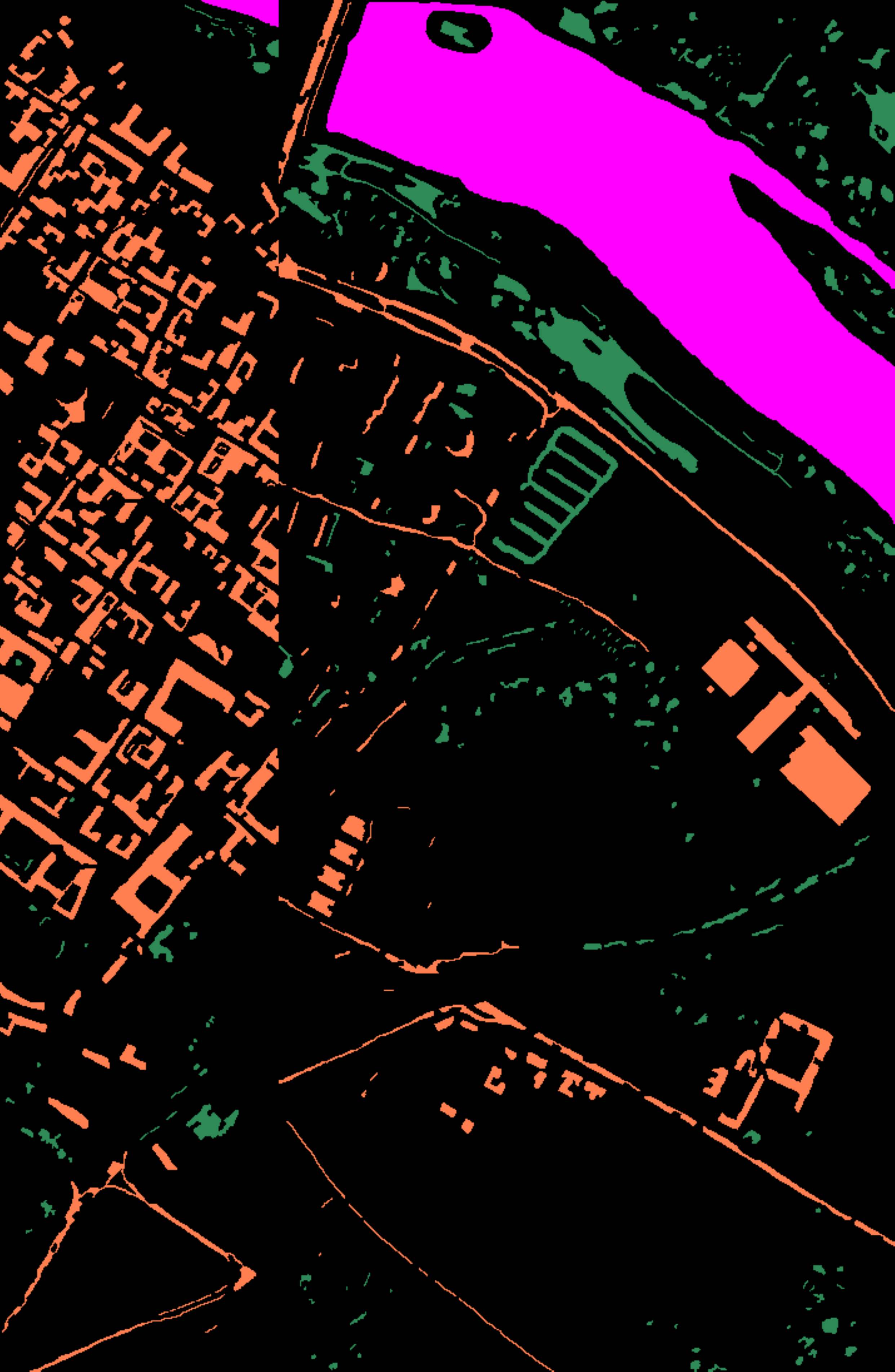}
	}
	\subfloat[]
	{
		\centering
		\includegraphics[scale=0.0625]{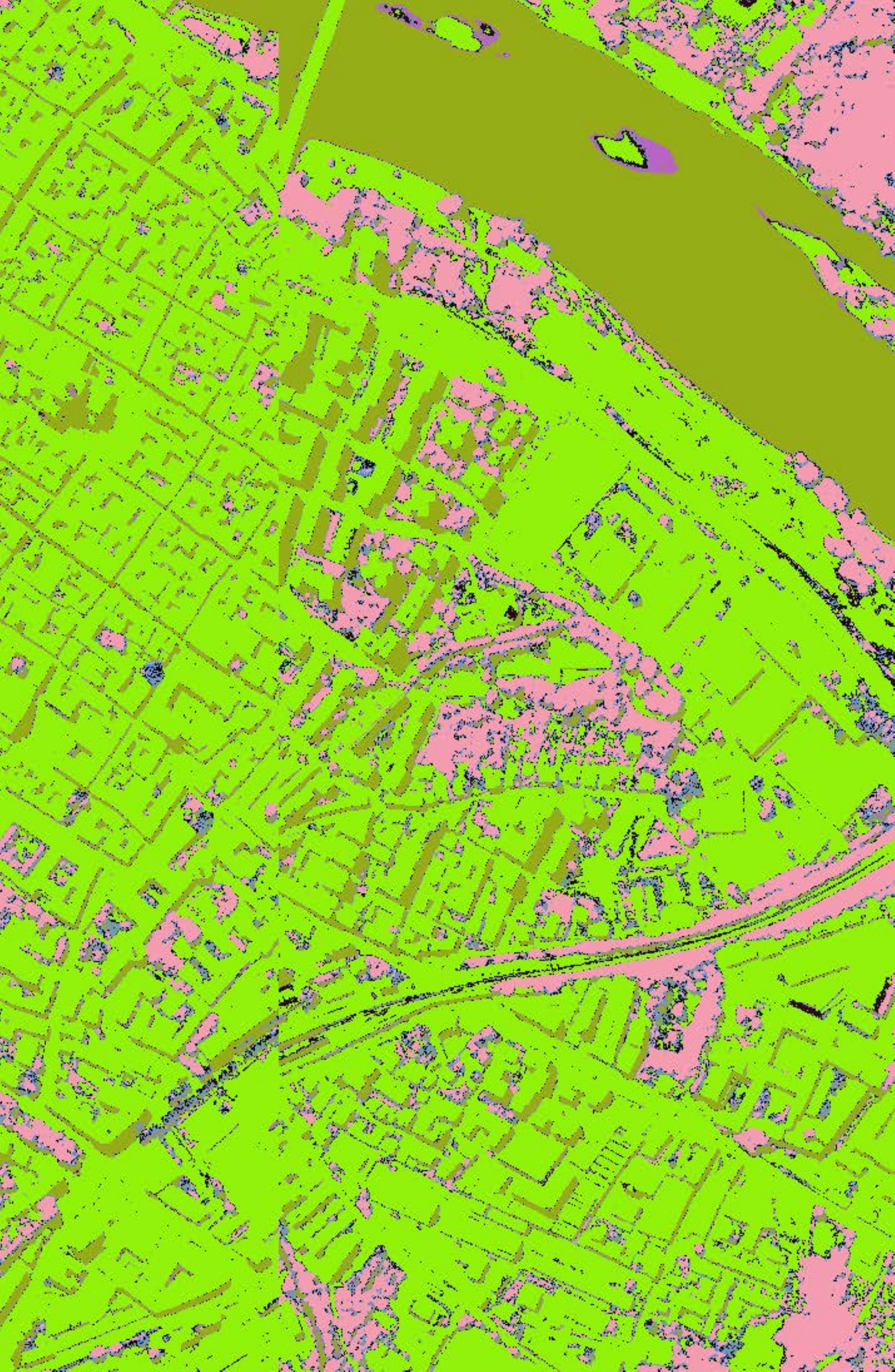}
	}
	\subfloat[]
	{
		\centering
		\includegraphics[scale=0.0625]{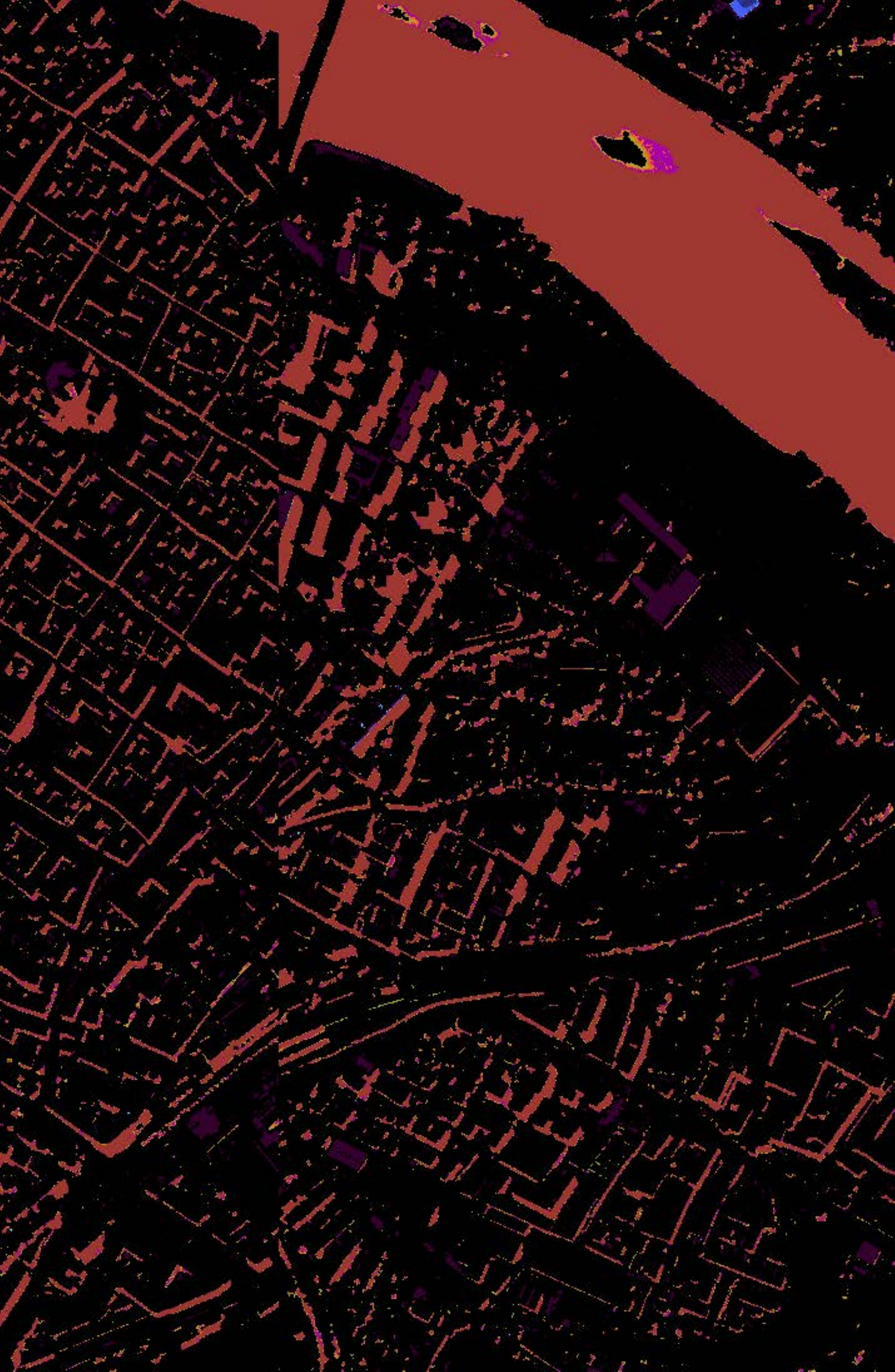}
	}
	\subfloat[]
	{
		\centering
		\includegraphics[scale=0.0625]{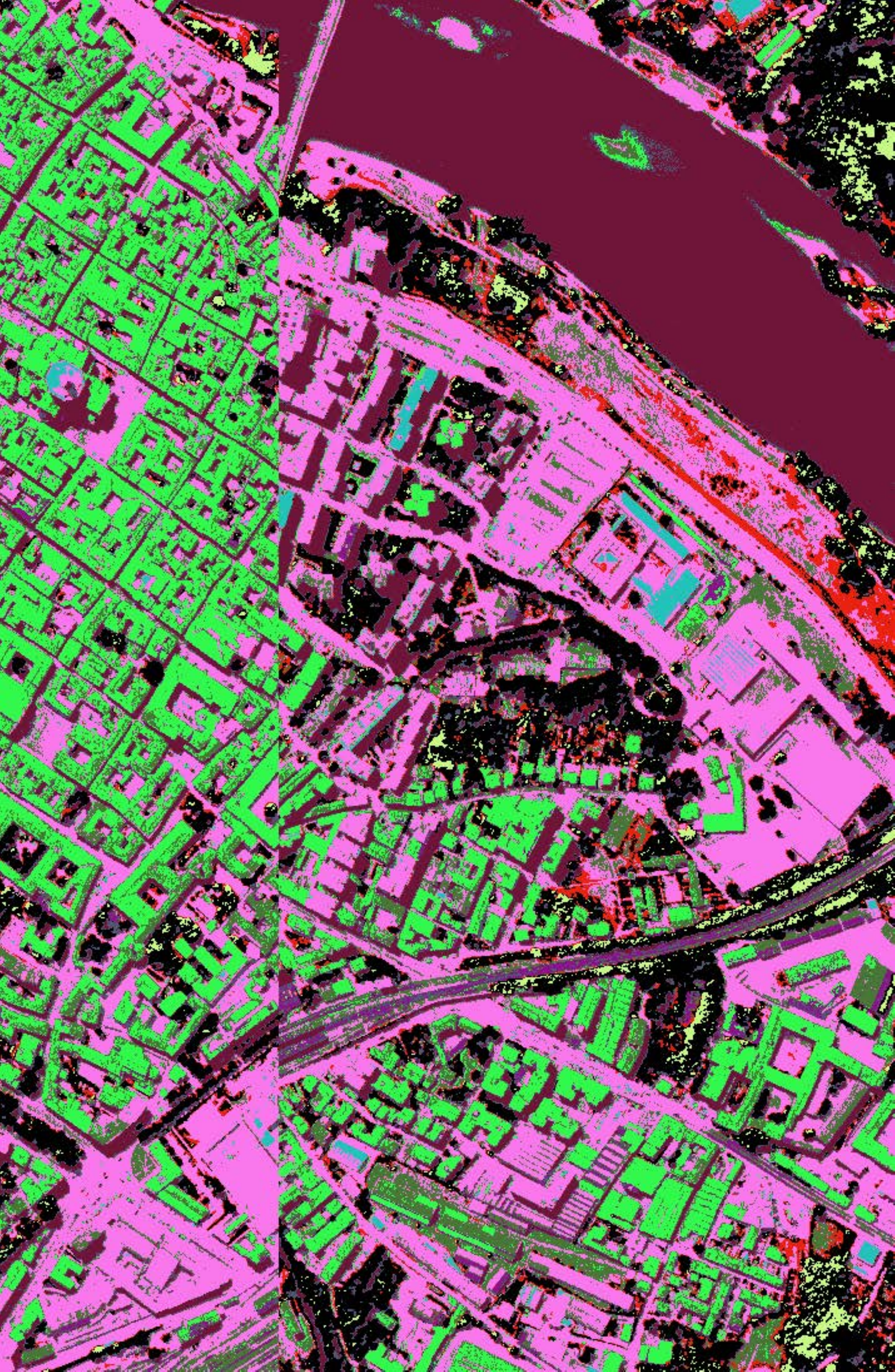}
	}\\
	\subfloat[]
	{
		\centering
		\includegraphics[scale=0.0625]{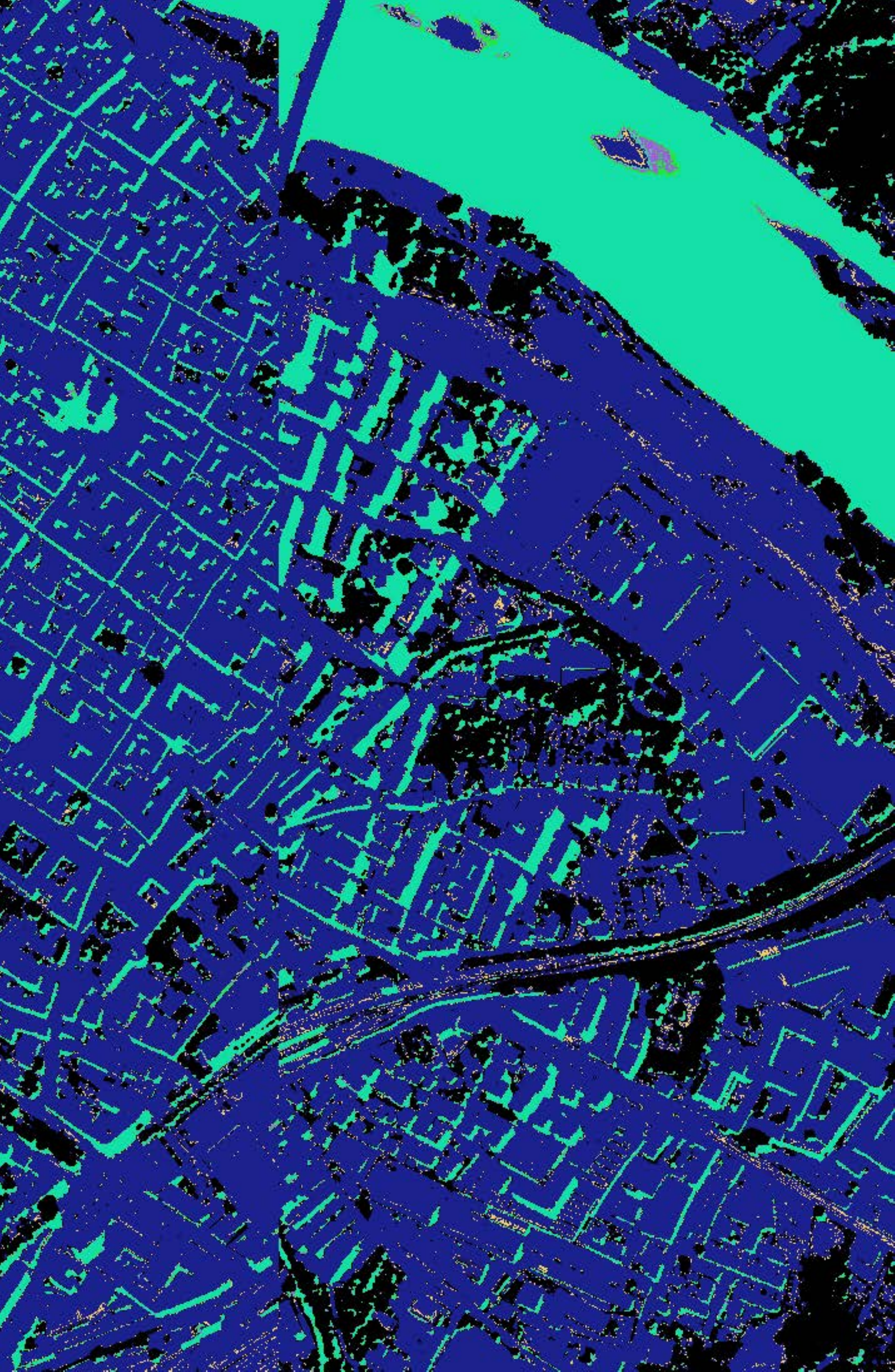}
	}
	\subfloat[]
	{
		\centering
		\includegraphics[scale=0.0625]{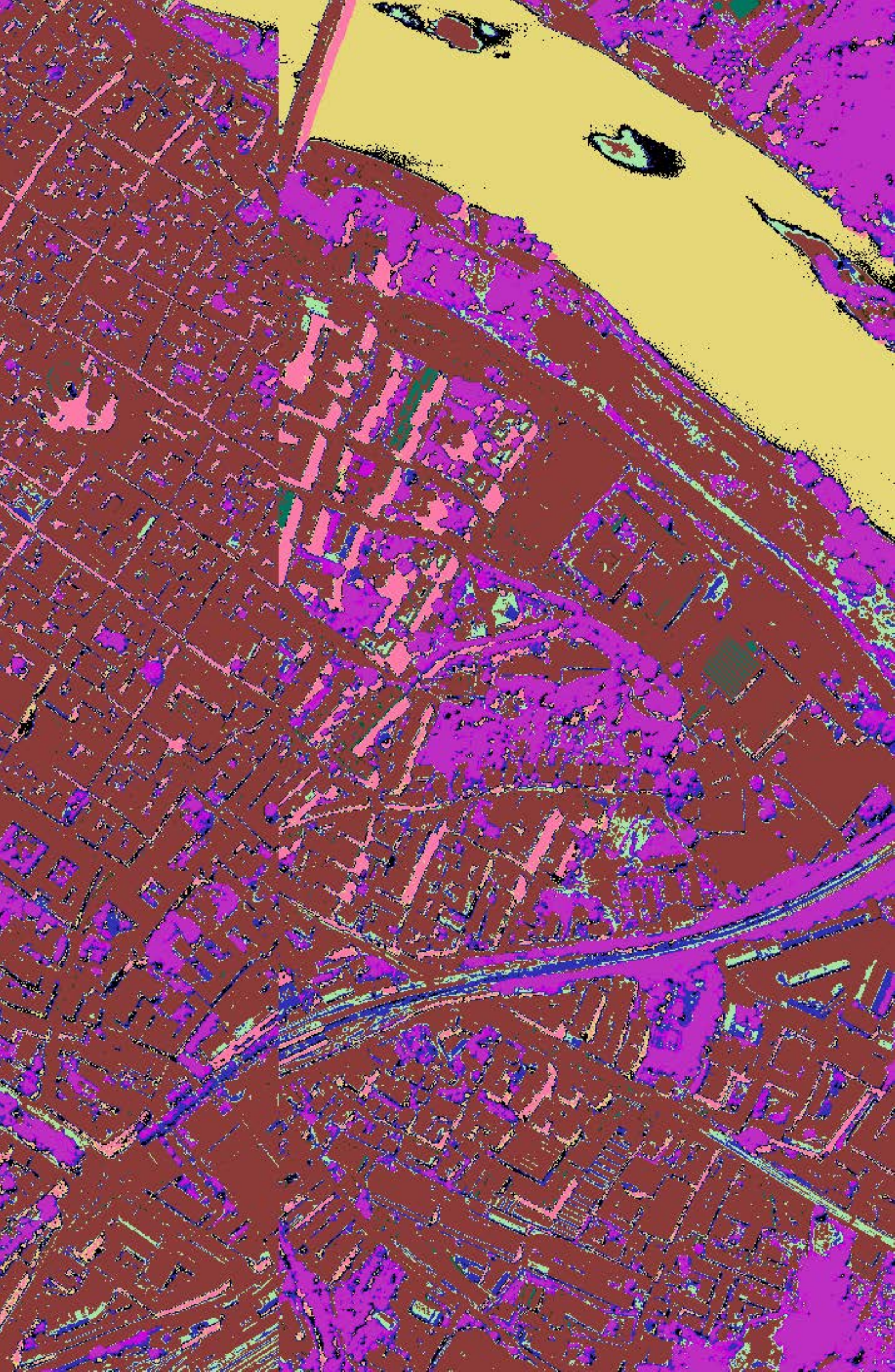}
	}
	\subfloat[]
	{
		\centering
		\includegraphics[scale=0.0625]{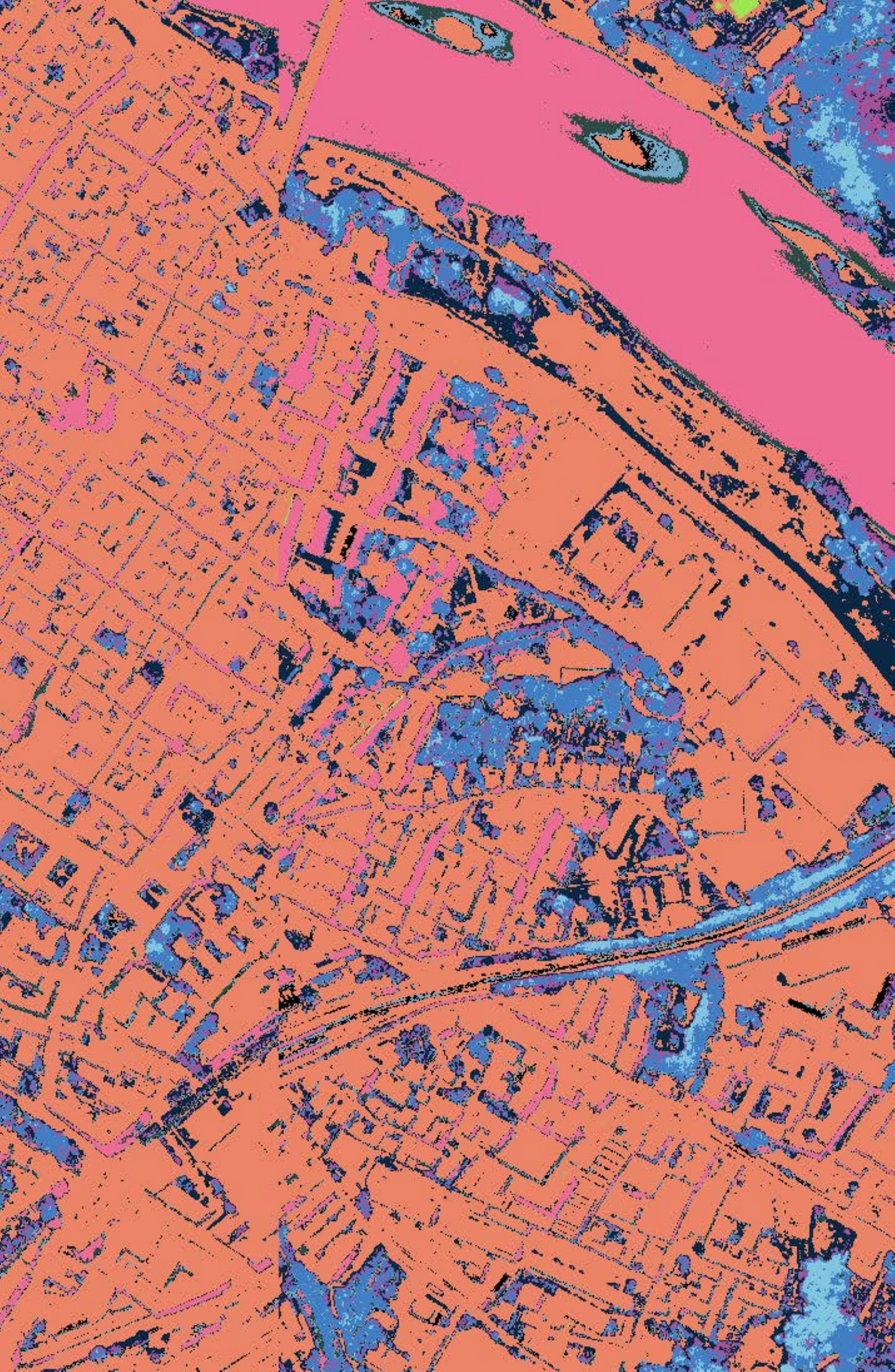}
	}
	\subfloat[]
	{
		\centering
		\includegraphics[scale=0.0625]{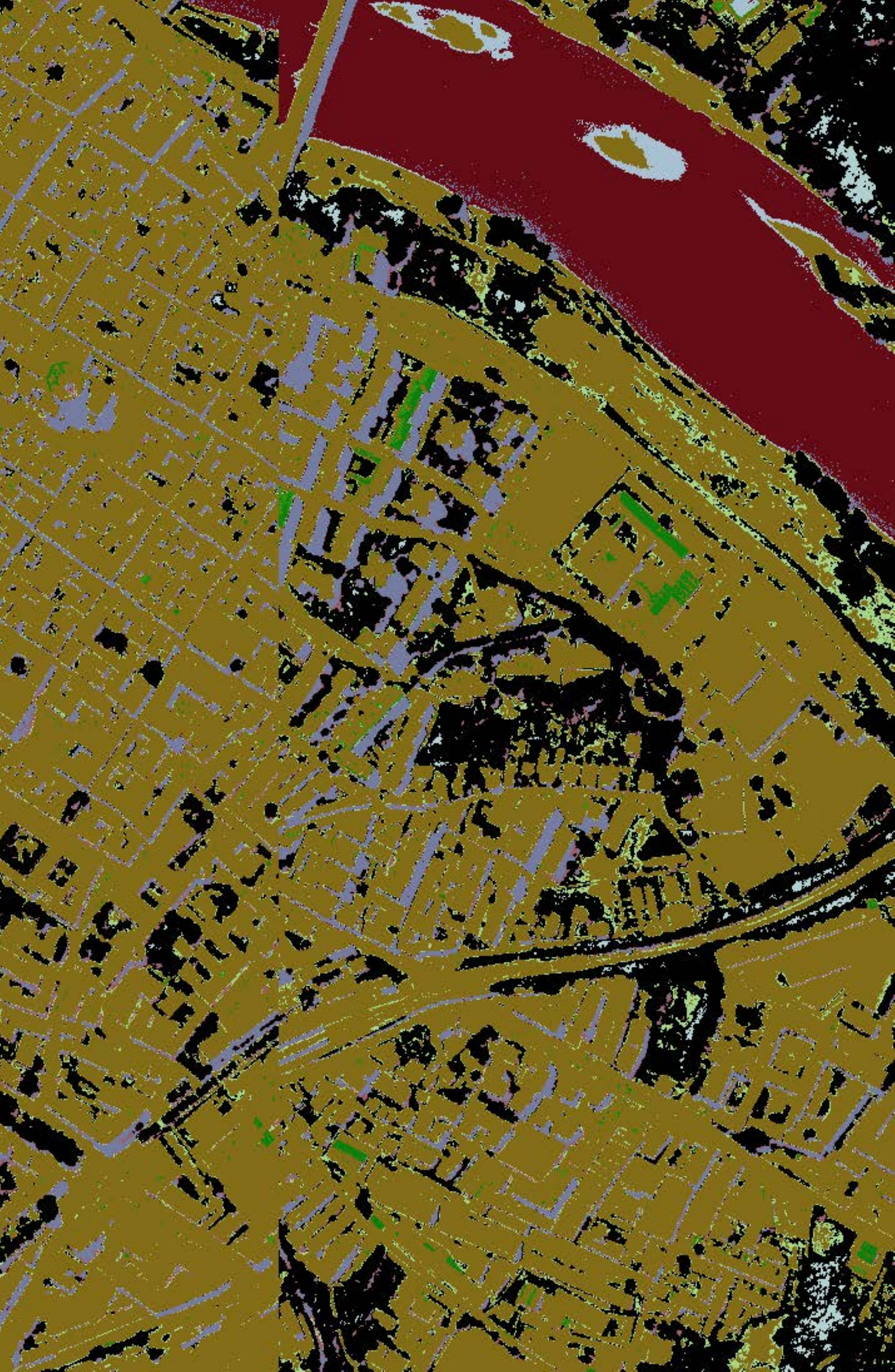}
	}
	\subfloat[]
	{
		\centering
		\includegraphics[scale=0.0625]{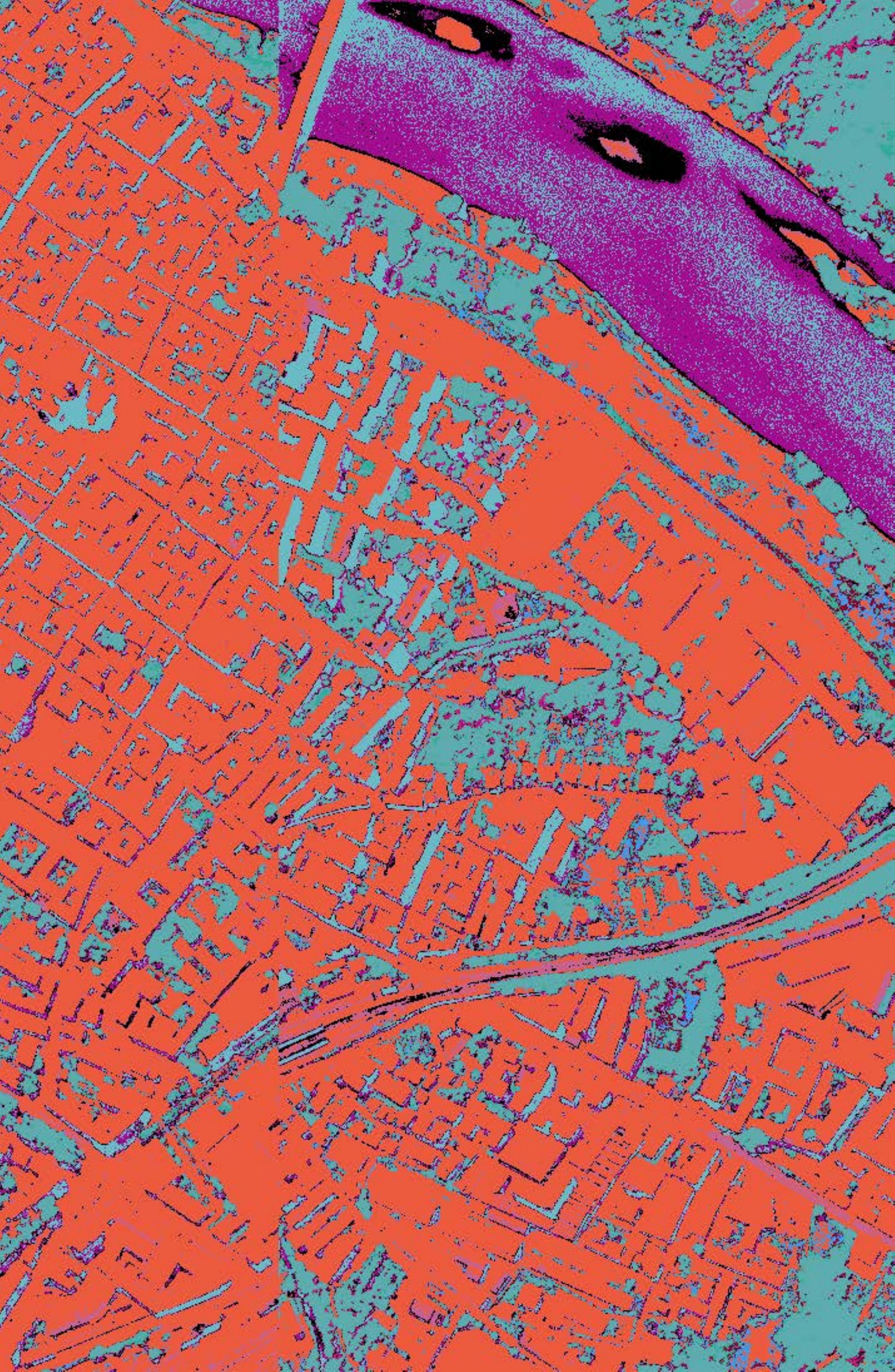}
	}\\
	\caption{The performance comparison of difierent algorithms on the dataset Pavia. (a) The Pavia image, (b) ground truth, (c) DCIDC, (d) PARTY, (e) AESSC, (f) SSC, (g) LS3C, (h) LRR, (i) LRSC, (j) SMR.}\label{result-pavia}
\end{figure}

In this subsection, we evaluate the performance of DCIDC on the four datasets respectively, by comparing with the baseline algorithms. In Tab. \ref{result-table-1} and Tab. \ref{result-table-2}, the bold names denote that the corresponding approaches are deep models which are at state of the art, and the bold scores mean the best results in the tables. Tab. \ref{result-table-1} and Tab. \ref{result-table-2} quantitatively describe the clustering accuracies and NMIs of DCIDC on datasets Indian Pines, Salinas, Salinas-A and Pavia. The experimental result shows that the proposed DCIDC has a better performance. The accuracies of DCIDC are at least $4.46\%$, $5.06\%$ and $1.94\%$ higher than that of the other methods regarding Indian Pines, Salinas, and Salinas-A datasets. For dataset Pavia, the accuracy of DCIDC is at least $1.24\%$ higher than the other methods. Fig. \ref{result-inidan-pines} -- \ref{result-pavia} qualitatively demonstrate the experimental results of the proposed DCIDC algorithm compared with other algorithms in a visual way.
\begin{table}[!htbp]
	\centering
	\caption{Performance comparison on Indian Pines and Salinas}\label{result-table-1}
	\begin{tabular}{c|cc|cc}
		\toprule
		Dataset&\multicolumn{2}{c|}{Indian Pines}&\multicolumn{2}{c}{Salinas}\\
		\midrule
		Method&Accuracy&NMI&Accuracy&NMI\\
		\toprule
		\textbf{DCIDC}&\textbf{89.22}&\textbf{93.78}&\textbf{90.56}&\textbf{93.42}\\
		\textbf{PARTY}&85.76&91.19&85.50&91.19\\
		\textbf{AESSC}&87.11&89.90&84.13&89.09\\
		\midrule
		SSC&84.30&91.04&81.08&90.12\\
		LS3C&30.98&49.23&30.37&40.50\\
		LRR&79.08&89.72&58.42&76.91\\
		LRSC&71.15&78.38&44.03&57.23\\
		SMR&80.49&89.42&74.86&84.47\\
		\bottomrule
	\end{tabular}
\end{table}
\begin{table}[!htbp]
	\centering
	\caption{Performance comparison on Salinas-A and Pavia}\label{result-table-2}
	\begin{tabular}{c|cc|cc}
		\toprule
		Dataset&\multicolumn{2}{c|}{Salinas-A}&\multicolumn{2}{c}{Pavia}\\
		\midrule
		Method&Accuracy&NMI&Accuracy&NMI\\
		\toprule
		\textbf{DCIDC}&\textbf{94.02}&\textbf{98.43}&\textbf{89.79}&\textbf{92.79}\\
		\textbf{PARTY}&92.08&96.91&88.55&90.85\\
		\textbf{AESSC}&88.81&94.43&74.80&78.33\\
		\midrule
		SSC&85.10&92.82&83.77&90.02\\
		LS3C&49.11&53.53&49.90&59.84\\
		LRR&81.00&93.01&81.62&89.16\\
		LRSC&68.68&73.20&68.29&73.40\\
		SMR&87.94&92.76&81.49&85.29\\
		\bottomrule
	\end{tabular}
\end{table}

\begin{figure}[!htbp]
	\centering
	\subfloat[]
	{
		\centering
		\includegraphics[scale=0.3]{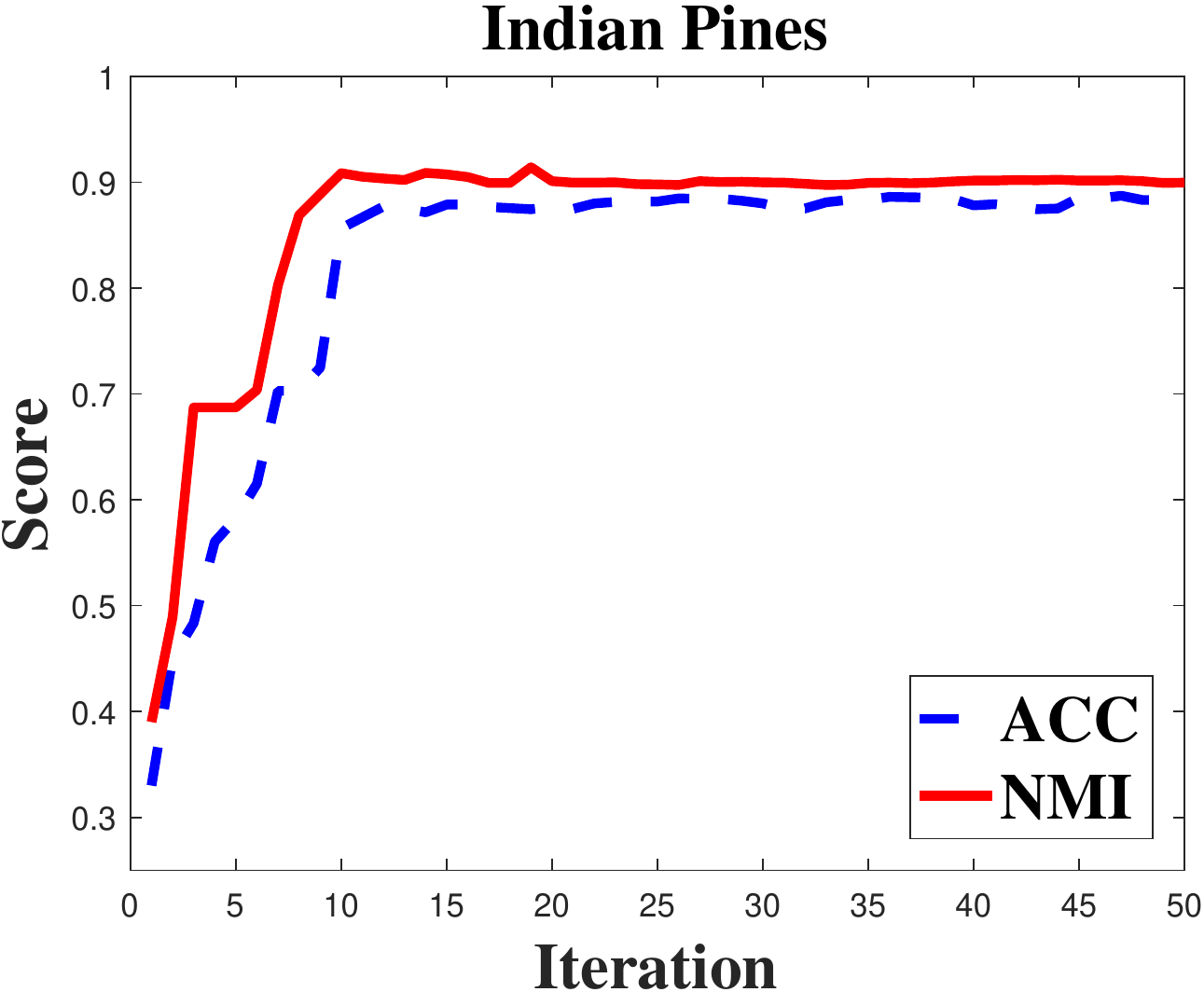}
	}
	\subfloat[]
	{
		\centering
		\includegraphics[scale=0.3]{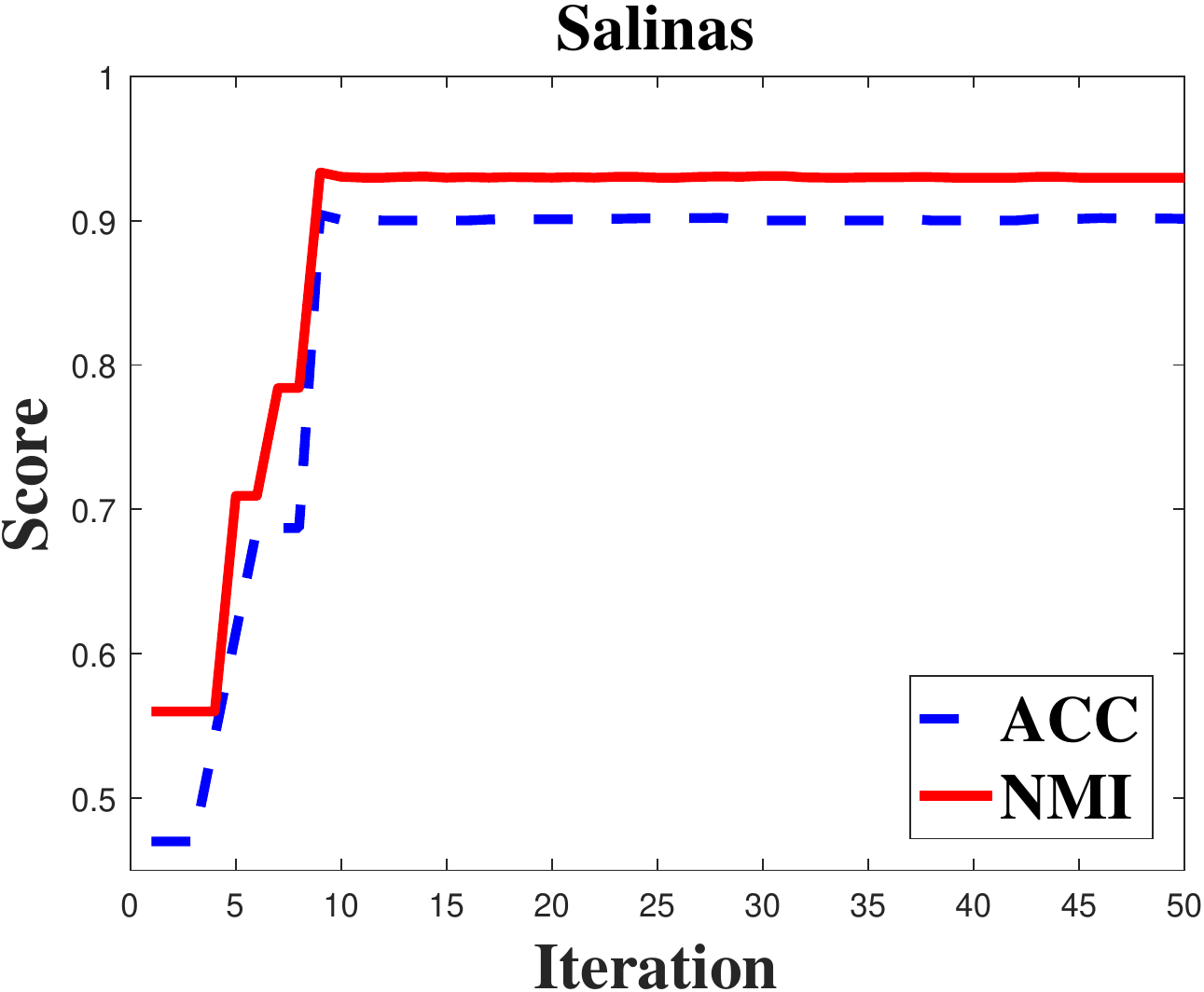}
	}\\
	\subfloat[]
	{
		\centering
		\includegraphics[scale=0.3]{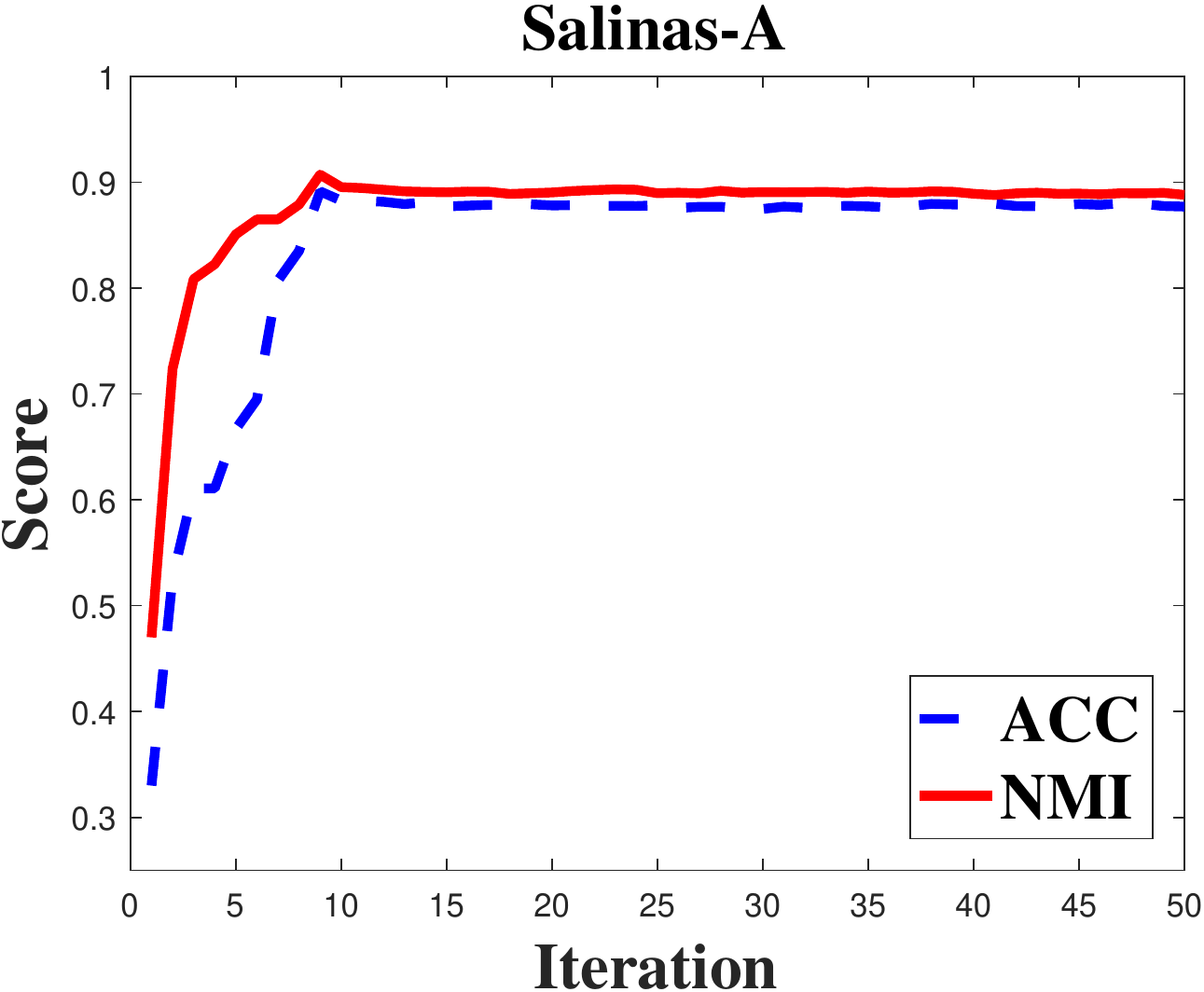}
	}
	\subfloat[]
	{
		\centering
		\includegraphics[scale=0.3]{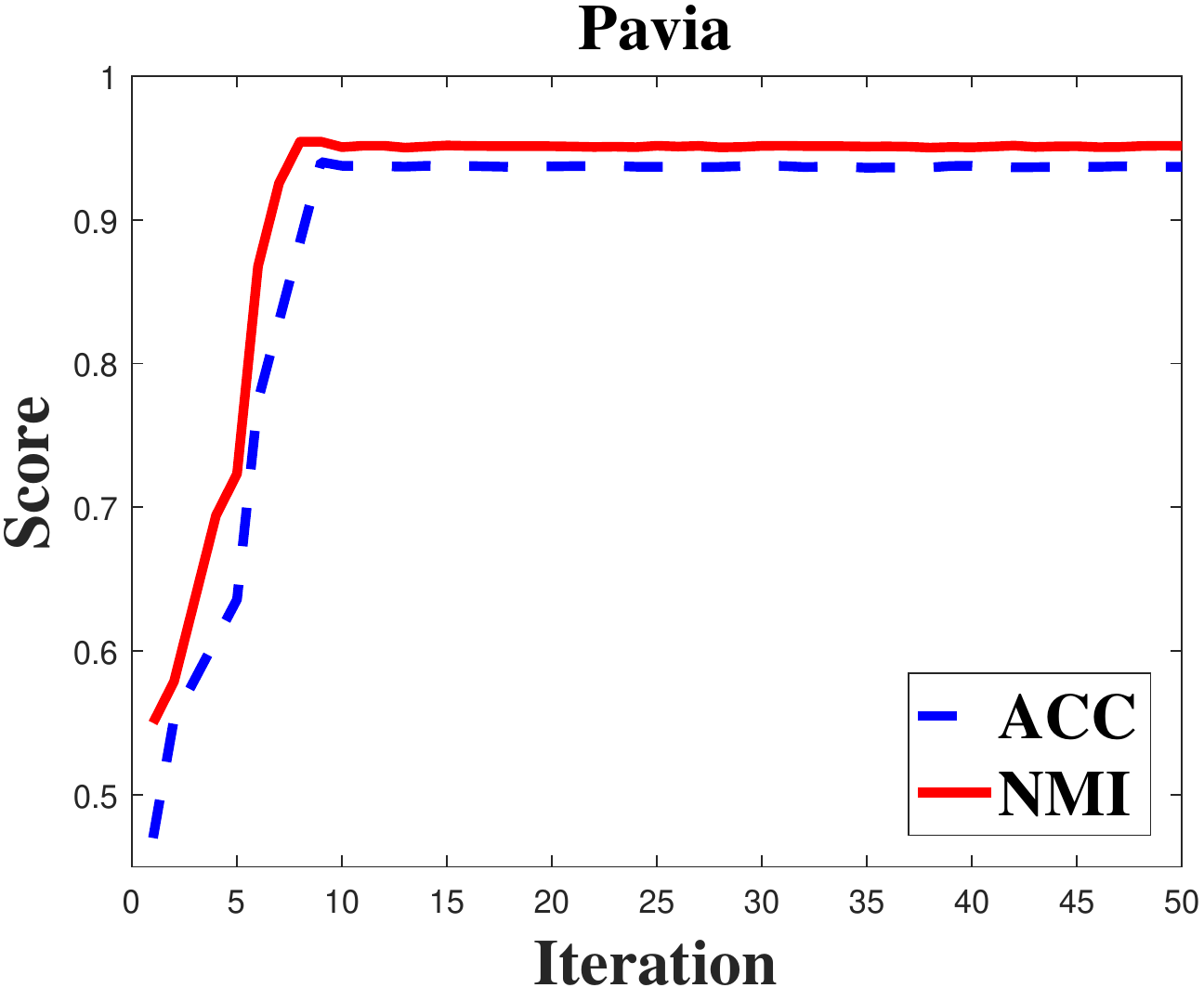}
	}\\
	\caption{The variation of Accuracy And NMI with iteration numbers in DCIDC. (a) Indian Pines, (b) Salinas, (c) Salinas-A, (d) Pavia.}\label{perform-iteration}
\end{figure}

In most cases, these results demonstrate that deep clustering methods perform much better than the shallow ones, benefiting from non-linear transformation and deep feature representation learning. However, the experimental results of the PARTY and AESSC methods do not outperform other shallow model-based algorithms. This may be due to that the auto-encoder based clustering methods mainly consider the reconstruction of input data, while the representation of the data lacks constraints of the prior knowledge. Our proposed DCIDC approach solves this issue by embedding intra-class distance into feature mapping process as a constraint, which can achieve higher accuracy. Fig. \ref{perform-iteration} shows the variety of accuracy and NMI in DCIDC as the iteration number increases on the four databases. It can be found that the performance is enhanced very fast in the first ten iterations, which implies that our method is effective and efficient. After dozens of iteration, both the accuracy and the NMI become stable, which shows that the proposed DCIDC approach is convergent.

\subsection{Influence of Tradeoff Coefficient}
\begin{figure}[!htbp]
	\centering
	\subfloat[]
	{
		\centering
		\includegraphics[scale=0.3]{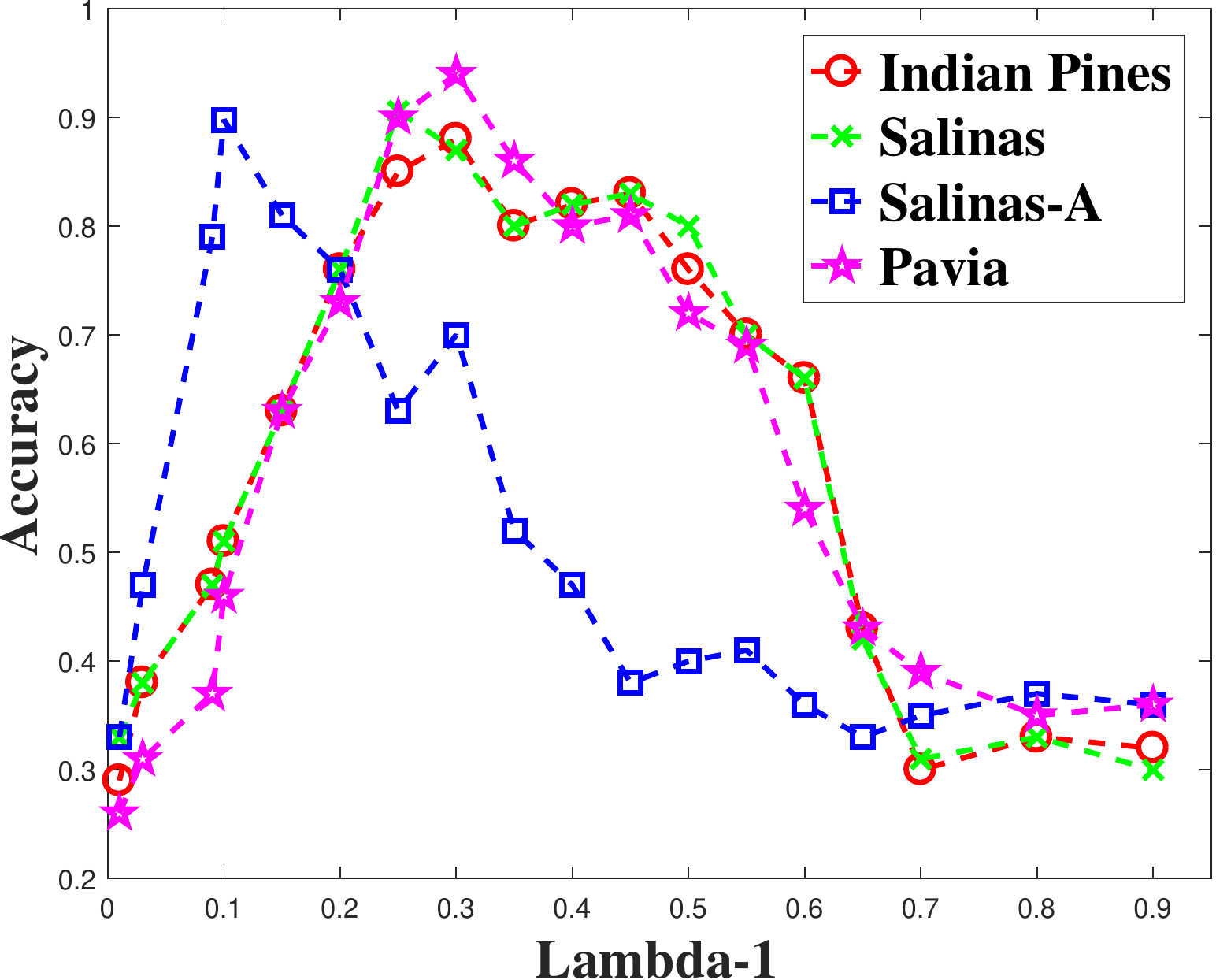}
	}
	\subfloat[]
	{
		\centering
		\includegraphics[scale=0.3]{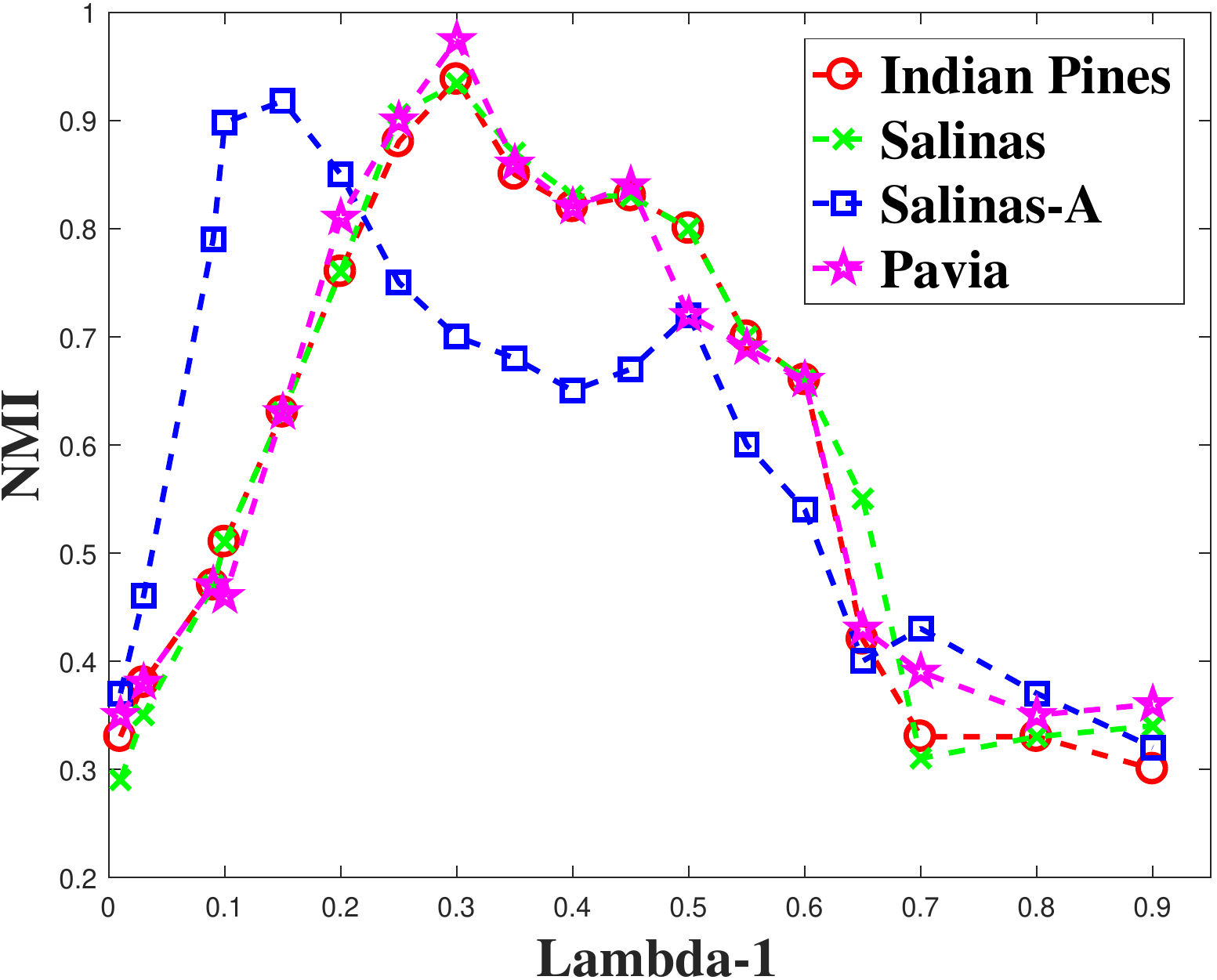}
	}
	\caption{The variation of Accuracy And NMI With $\lambda_1$ in DCIDC. (a) Accuracy, (b) NMI.}\label{variation-lambda}
\end{figure}

The coefficient $\lambda_2$ is designed to prevent over-fitting, and its impact on DCIDC is balanced in different scenarios, so we set it as a fixed value of 0.0003. For the trade-off coefficient $\lambda_1$, we investigate the variation of accuracy and NMI with different values of $\lambda_1$ to obtain the optimal value of $\lambda_1$. The variations of accuracy and NMI with different values of $\lambda_1$  are shown in Fig. \ref{variation-lambda}. It can be found that given different $\lambda_1$ values, the model achieves different clustering accuracies and NMIs. In most cases, the optimal value of $\lambda_1$ is near 0.3.

\subsection{Influence of Activation Functions}
\begin{figure}[!htbp]
	\centering
	\subfloat[]
	{
		\centering
		\includegraphics[scale=0.25]{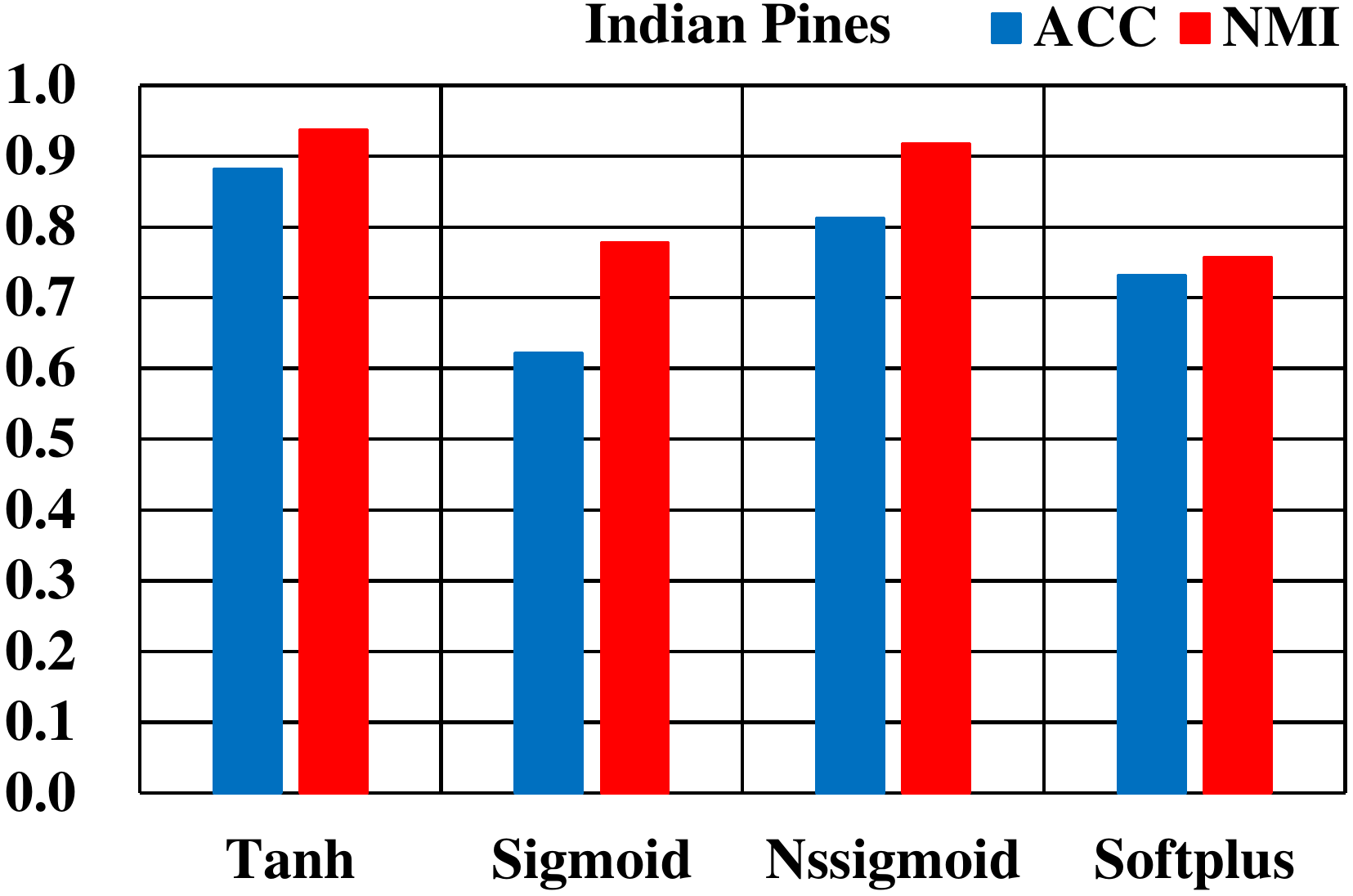}
	}
	\subfloat[]
	{
		\centering
		\includegraphics[scale=0.25]{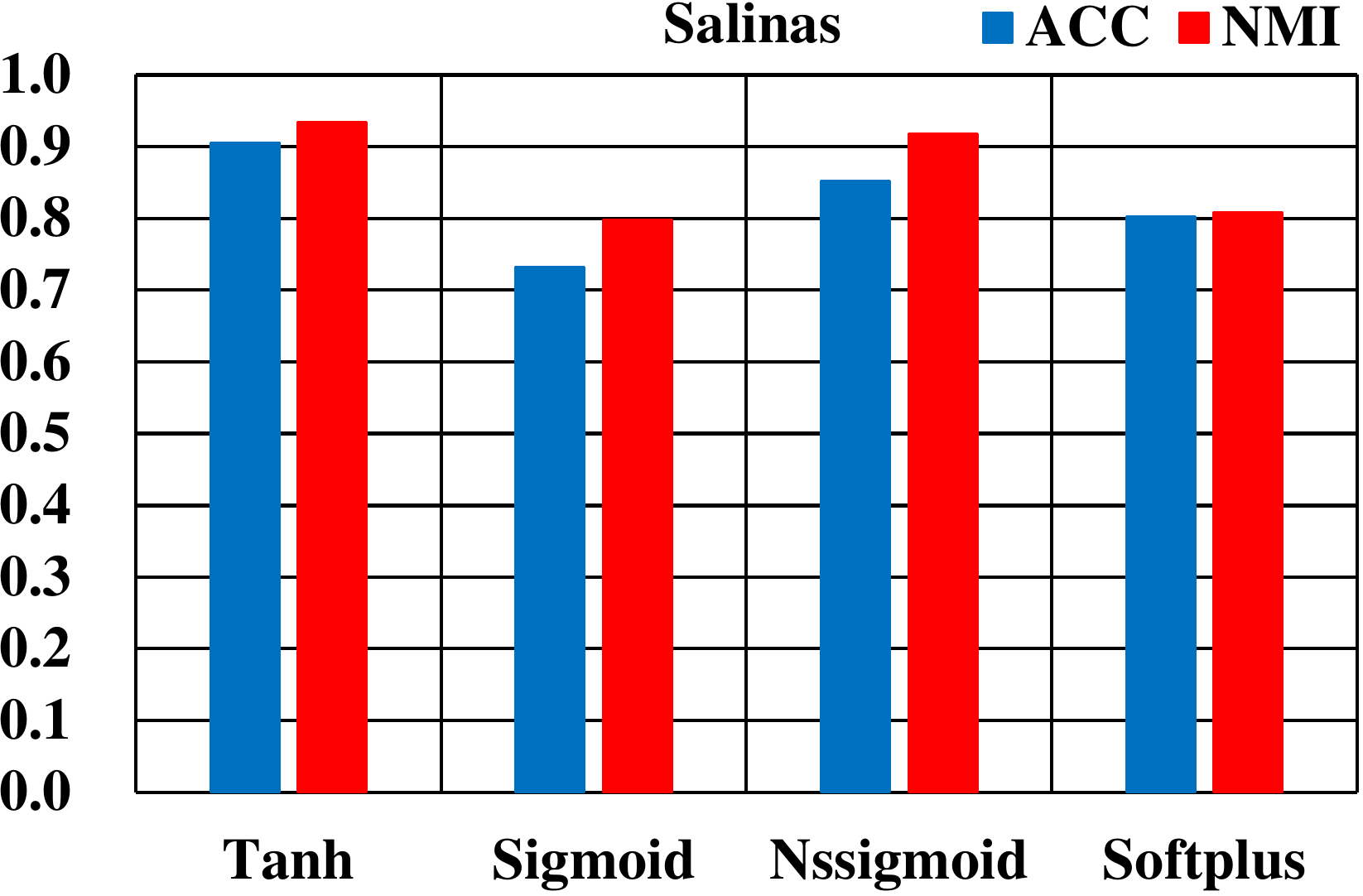}
	}\\
	\subfloat[]
	{
		\centering
		\includegraphics[scale=0.25]{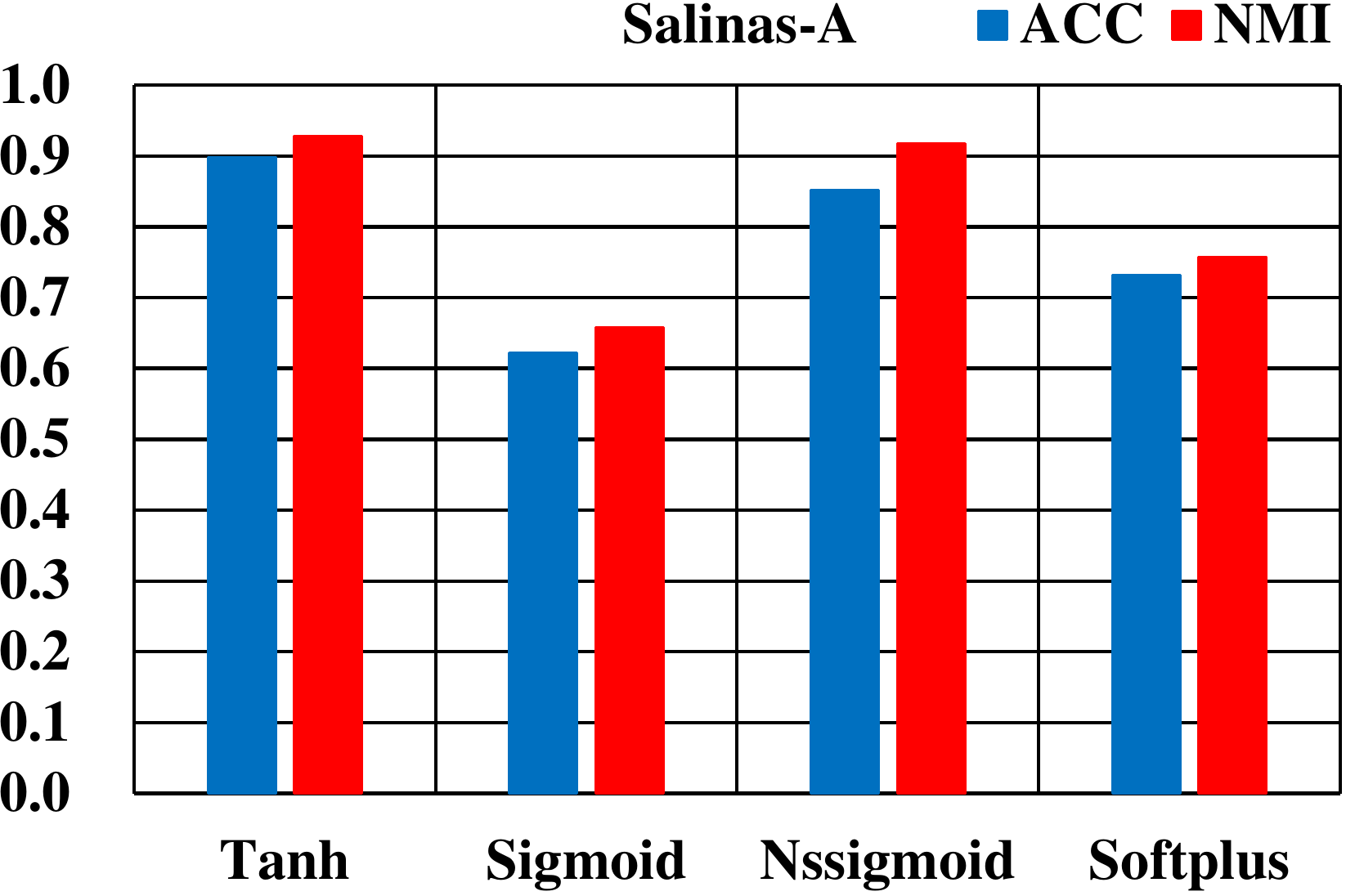}
	}
	\subfloat[]
	{
		\centering
		\includegraphics[scale=0.25]{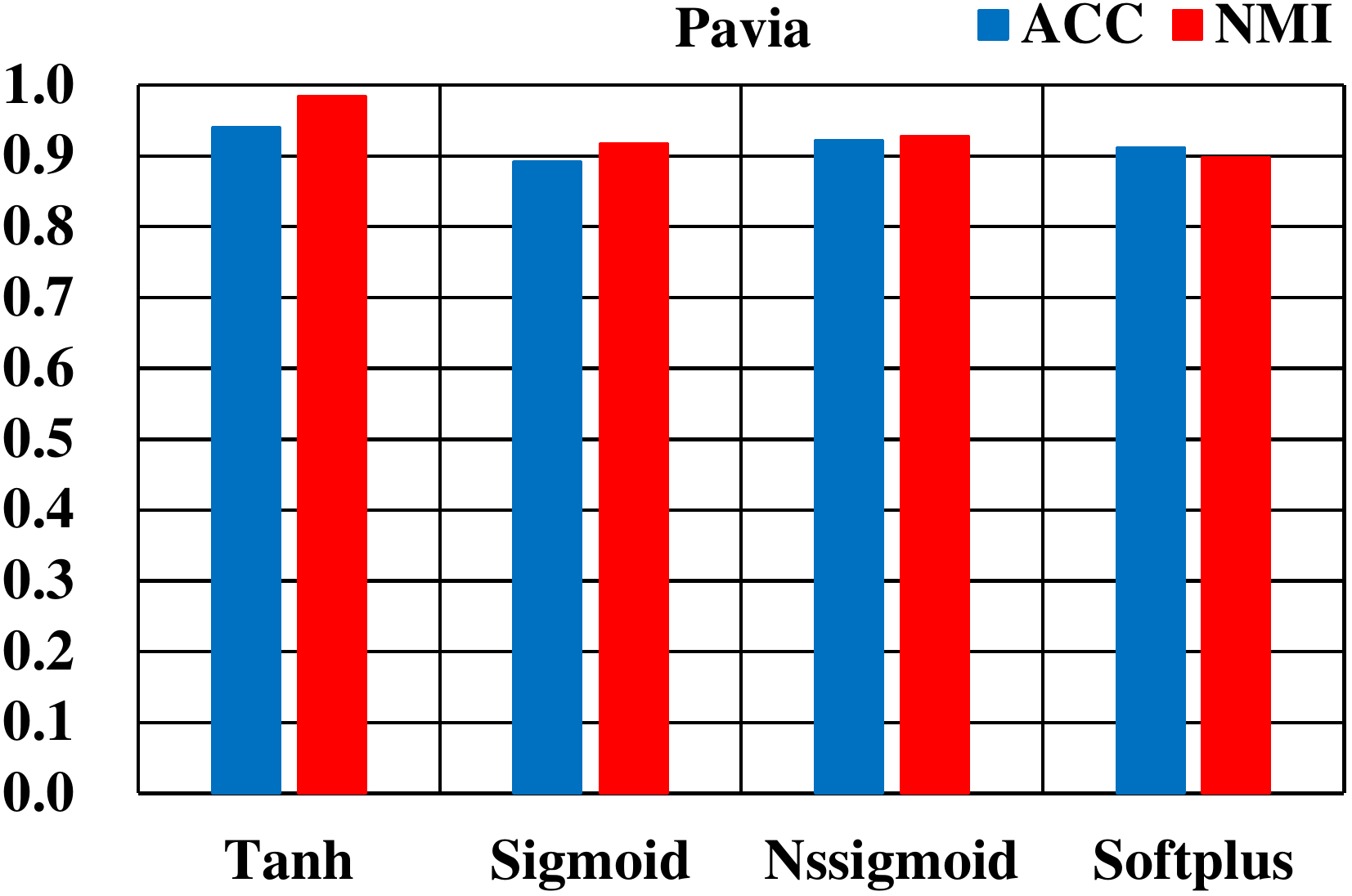}
	}\\
	\caption{The performance of DCIDC With four different activation functions. (a) Indian Pines, (b) Salinas, (c) Salinas-A, (d) Pavia.}\label{variation-functions}
\end{figure}

In this subsection, we report the performance of DCIDC with four different activation functions on the four datasets. The used activation functions includes \emph{Tanh}, \emph{Sigmoid}, \emph{Nssigmoid}, and \emph{Softplus}. From Fig. \ref{variation-functions}, we can see that the \emph{Tanh} function outperforms the other three activation functions in the experiments and the \emph{Nssigmoid} function achieves the second best result which is very close to the best one.

\section{Conclusions}

In this paper, we propose an intra-class distance constrained deep clustering approach, which constrains the feature mapping procedure by intra-class distance. Compared with the current deep clustering approaches, the procedure of feature extraction of DCIDC is more purposeful, making the features learned more conducive to clustering. The proposed approach jointly optimize the parameters of the network and the procedure of the clustering without additional pre-training process, which is more efficient. We conducted comparative experiments on four different hyperspectral datasets, and the accuracies of DCIDC are at least $4.46\%$, $5.06\%$, $1.94\%$, and $1.24\%$ higher than that of the compared methods, regarding the datasets of Indian Pines, Salinas, Salinas-A ,and Pavia. The experimental results demonstrate that our approach remarkably outperforms the state-of-the-art clustering methods in terms of accuracy and NMI. In the future, the adaptive deep clustering methods will be further explored based on this work.


%








\bibliographystyle{IEEEtran}
\bibliography{Reference}
\end{document}